%% file: acl_latex.tex
\documentclass[11pt]{article}

\usepackage[preprint]{acl}

\usepackage{times}
\usepackage{latexsym}

\usepackage[T1]{fontenc}

\usepackage[utf8]{inputenc}

\usepackage{microtype}

\usepackage{inconsolata}

\usepackage{graphicx}

\usepackage{hyperref}       
\usepackage{url}            
\usepackage{booktabs}       
\usepackage{amsfonts}       
\usepackage{nicefrac}       
\usepackage{xcolor}         
\usepackage[most]{tcolorbox}
\usepackage{cleveref} 
\usepackage{multirow}
\usepackage{placeins}
\usepackage{colortbl}
\usepackage{siunitx}
\usepackage{xspace}
\usepackage{caption}
\usepackage{bm}
\usepackage{amssymb} 
\usepackage{subcaption}

\newcommand{\name}[1]{\textsc{HEDGE}\xspace}

\newcommand{\score}[2]{%
    \ensuremath{#1_{\textcolor{red}{\scriptscriptstyle\pm #2}}}%
}
\newcommand{\bestscore}[2]{%
    \ensuremath{\mathbf{#1}_{\textcolor{red}{\scriptscriptstyle\pm #2}}}%
}

\title{\name{}: Gated Hypergraph Memory for Scalable Multi-Agent Debate}

\author{
\textbf{
Amirhossein Abaskohi$^{1}$\thanks{Corresponding author: \texttt{aabaskoh@cs.ubc.ca}.},
Mahdi Mostajabdaveh$^{2}$,
Zirui Zhou$^{2}$}\\[0.5em]
$^{1}$University of British Columbia, $^{2}$Huawei Technologies Canada
}

\begin{document}


\newcounter{prompt}[section]
\renewcommand{\theprompt}{\thesection.\arabic{prompt}}

\newenvironment{promptbox}[2][]{%
    \refstepcounter{prompt}%
    \begin{tcolorbox}[
        colback=blue!5!white,
        colframe=blue!75!black,
        fonttitle=\bfseries,
        title={Prompt \theprompt: #2},
        enhanced jigsaw,
        breakable,
        attach boxed title to top left={
            yshift=-2mm,
            xshift=5mm
        },
        boxed title style={
            colback=blue!75!black
        },
        sharp corners=south,
        drop shadow,
        left=1mm,
        right=1mm,
        top=2mm,
        bottom=1mm,
        boxsep=1mm,
        #1
    ]%
    \ttfamily\normalsize
    \linespread{1.2}\selectfont
    \obeylines
    \obeyspaces
    \ignorespaces
}{%
    \end{tcolorbox}
}

\newenvironment{promptbox*}[2][]{%
    \begin{figure*}[t]
    \refstepcounter{prompt}%
    \begin{tcolorbox}[
        colback=blue!5!white,
        colframe=blue!75!black,
        fonttitle=\bfseries,
        title={Prompt \theprompt: #2},
        enhanced,
        width=\textwidth,
        attach boxed title to top left={
            yshift=-2mm,
            xshift=5mm
        },
        boxed title style={
            colback=blue!75!black
        },
        sharp corners=south,
        drop shadow,
        left=1mm,
        right=1mm,
        top=2mm,
        bottom=1mm,
        boxsep=1mm,
        #1
    ]%
    \ttfamily\small
    \linespread{1.2}\selectfont
    \obeylines
    \obeyspaces
    \ignorespaces
}{%
    \end{tcolorbox}
    \end{figure*}
}

\maketitle


\input{sections/1_abstract}
\input{sections/2_introduction}
\input{sections/3_related_work}
\input{sections/4_memory}

\input{sections/5_msm}
\input{sections/6_experiments}

\input{sections/7_conclusion}


\section*{Limitations}
\name{} has some limitations. First, MSM is trained using pseudo-labels derived from debate trajectories rather than human-annotated memory decisions, so it may discard claims whose utility becomes apparent only in later rounds. Second, memory construction depends on accurately decomposing agent responses into atomic items and representing them through embeddings and predicted relations; errors in either step may propagate through the hypergraph. Third, Stage~II relies on reward weights and routing thresholds that balance task accuracy against memory compactness, and different settings may shift this trade-off.

\section*{Ethical Considerations}
\label{sec:ethics}

\paragraph{Potential risks.}
\name{} inherits the limitations and risks of its underlying language models, including hallucinations, biases, and unsupported reasoning. Because information is shared across agents, an incorrect claim admitted to memory may influence later rounds or create false consensus. Errors in decomposing responses or predicting intra-hyperedge relations may similarly distort the stored argumentative structure. Although MSM is designed to filter low-utility content, it does not guarantee factuality or fairness. Deployments in high-stakes domains should therefore include source verification, human oversight, and auditing of both model outputs and stored memory.


\paragraph{Computational cost.}
Multi-agent debate remains more computationally expensive than single-agent inference. While \name{} reduces token usage, memory footprint, and debate rounds relative to strong memory-based systems, its overall computational and environmental costs remain non-trivial. We report detailed efficiency measurements in Appendix~\ref{app:detailed_results_efficiency} to support assessment of this trade-off.

\paragraph{Use of AI assistance.}
During manuscript preparation, we used large language models to improve phrasing and presentation and to generate icons used in figures. All technical contributions, experimental designs, analyses, and conclusions were developed and verified by the authors.

\paragraph{Reproducibility statement.}
We describe the memory construction procedure, MSM architecture, training objectives, hyperparameters, and evaluation protocol in the main paper and appendices. The MSM data-generation pipeline, including debate simulation, memory logging, and pseudo-label construction, is detailed in Appendix~\ref{sec:msm_data_generation}, and all prompts are provided in Appendix~\ref{app:prompts}. Upon publication, we will release the implementation, MSM training scripts, generated trajectories, checkpoints, and evaluation code under a permissive research license.

\paragraph{Licensing and artifact release.}
We use external datasets, models, and baseline implementations according to their original licenses and terms of use. Before release, we will document the source, license, and required attribution for each dependency. Our original code will be released under a permissive license. Generated trajectories and checkpoints will be redistributed only when permitted by the licenses and terms governing their source datasets and models; otherwise, we will release generation scripts, configuration files, and dataset identifiers needed to reproduce them. We will not redistribute third-party model weights, datasets, or baseline artifacts when their licenses prohibit doing so.

\paragraph{Use of existing artifacts and data.}
All experiments use publicly available benchmarks, models, and baseline implementations according to their original licenses and access conditions. MSM training trajectories are generated from the HotpotQA full-wiki training split~\citep{yang2018hotpotqa}, which is derived from Wikipedia. We do not collect new user data, attempt to identify individuals, or involve human subjects. Naturally occurring names or other entities may remain in the original benchmark content, and our released artifacts are intended solely for research and reproducibility.


\bibliography{custom}



\appendix

\input{sections/app_maintenance_details}
\input{sections/app_MSM_architecture}
\input{sections/app_data_generation}
\input{sections/app_MSM_training}
\input{sections/app_baseline_benchmarks}
\input{sections/app_MAS_frameworks}
\input{sections/app_experimental_settings}

\input{sections/app_qualitative}
\input{sections/app_generalization}
\input{sections/app_sensitivity_to_retrieval_breadth}
\input{sections/app_sensitivity_to_embedding_model}
\input{sections/app_sensitivity_extraction_model}

\input{sections/app_detailed_efficiency_results}
\input{sections/app_signature_importance}
\input{sections/app_prompts}

\end{document}

%% file: sections/1_abstract.tex
\begin{abstract}
Multi-agent debate systems require agents to retain and revisit arguments across multiple rounds, yet limited context windows and unstructured memories make relevant information increasingly difficult to recover. Existing flat and pairwise-graph memories also struggle to capture higher-order relations, including multi-agent agreement, conflicting evidence, and evolving consensus. We introduce \textbf{\name{}}, a selective memory framework that represents debate trajectories as semantic hypergraphs. Hyperedges connect related claims across agents, while intra-hyperedge relational graphs preserve fine-grained argumentative structure. A learned \emph{memory selection module} (MSM), trained through supervised learning and reinforcement learning, routes each memory candidate to \textsc{store}, \textsc{backup}, or \textsc{omit}, retaining useful information while limiting redundancy. Across seven diverse benchmarks, \name{} consistently outperforms strong single-agent, multi-agent, and memory baselines, improving accuracy by 5.0 points on average over the strongest multi-agent system with GPT-4o-mini. The gains reach 8.2 points on conflict-heavy tasks, while \name{} reduces memory usage by 48\% and the number of debate rounds by 34\%. Code is available at: \url{https://github.com/AmirAbaskohi/HEDGE}.
\end{abstract}

%% file: sections/2_introduction.tex
\section{Introduction}
\label{sec:introduction}

\begin{figure}[t]
    \centering
    \includegraphics[width=1\linewidth]{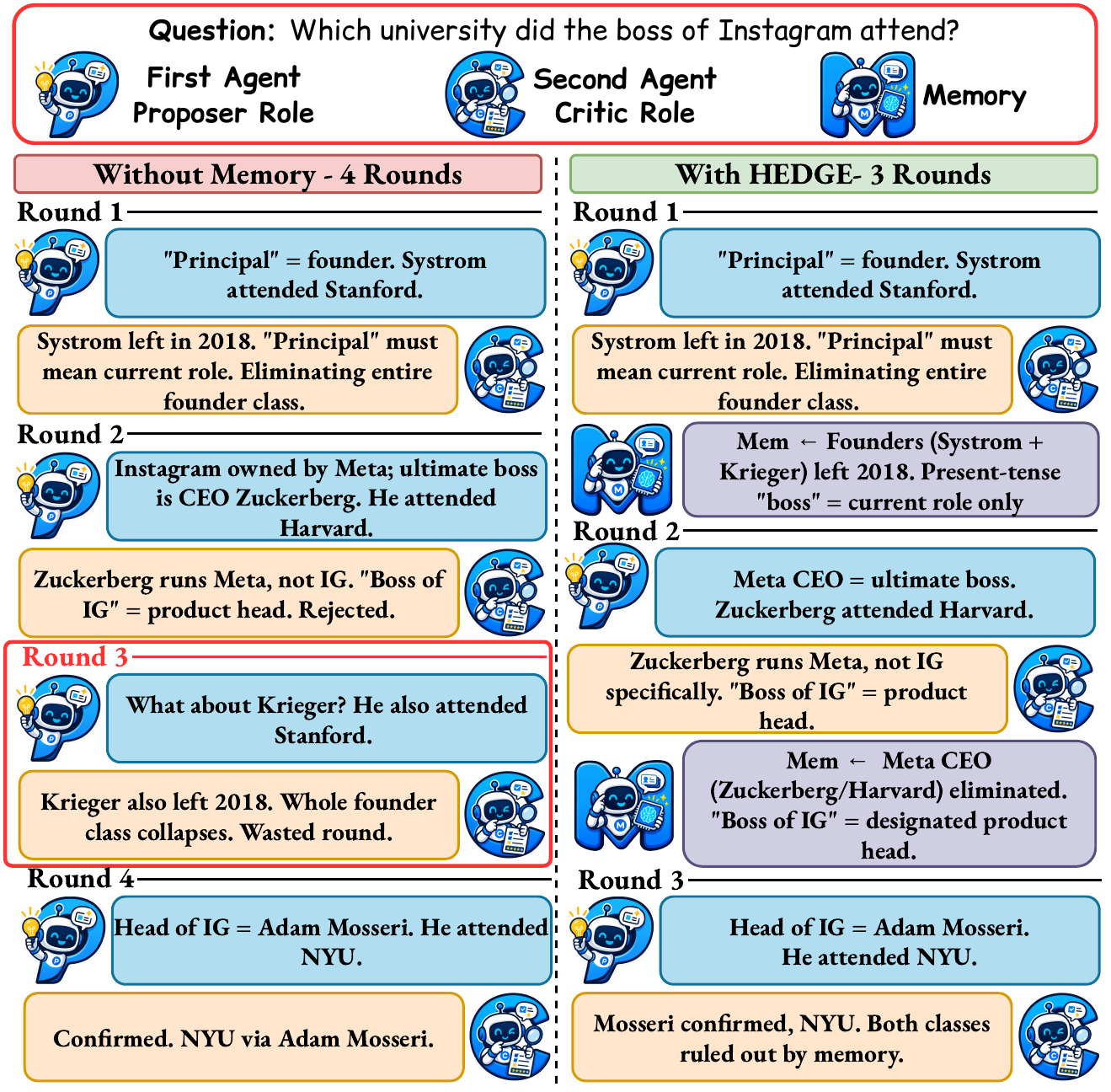}
    \caption{Trajectory-level memory in \name{} during multi-agent debate. Without memory (\textbf{\textit{left}}), agents revisit a rejected founder hypothesis; with \name{} (\textbf{\textit{right}}), the stored rationale avoids redundant reasoning and reaches the correct answer in 3 instead of 4 rounds.}
    \label{fig:teaser}
    \vspace{-1em}
\end{figure}

Multi-agent debate (MAD) frameworks, in which multiple LLM-based agents propose, critique, and refine arguments over several iterative rounds, have emerged as an effective inference-time strategy for complex reasoning~\citep{du2024improving,liang-etal-2024-encouraging}. By fostering divergent thinking, MAD mitigates the degeneration-of-thought problem, where a single model becomes trapped in a flawed reasoning path~\citep{liang-etal-2024-encouraging}, and produces more reliable consensus than single-agent evaluation~\citep{chan2024chateval,chen-etal-2024-reconcile}.

However, as debates grow longer, their expanding histories make earlier evidence, rejected hypotheses, and cross-round dependencies increasingly difficult to recover~\citep{liu-etal-2024-lost}. Important arguments may become buried in the transcript, causing agents to repeat previously rejected claims, overlook unresolved conflicts, or reconsider evidence without its original context. Retaining the full interaction history is equally problematic, as irrelevant and redundant content increases token cost and can degrade reasoning quality~\citep{jiang2026anatomy,yu2026memagent}. Recent methods address parts of this challenge: MAD-M$^2$ masks erroneous memories between rounds~\citep{tian2026multiagentdebatememorymasking}, while EquiMem calibrates updates to shared memory~\citep{meng2026equimemcalibratingsharedmemory}. However, neither jointly organizes debate trajectories into structured argumentative memory, learns the evolving utility of individual memory items, and supports continuous access throughout the debate.

Simply pruning the debate history using heuristic rules is insufficient because an argument's utility changes as the interaction evolves~\citep{yan2025memory}. Effective debate memory therefore requires a \emph{learned} mechanism that determines what to retain, back up, or discard. It also requires an appropriate structure: flat text or pairwise graphs~\citep{zhang2025gmemory} cannot naturally represent higher-order relations such as multi-agent agreement, conflicting evidence chains, and evolving consensus. Finally, memory must be integrated into the intra-debate reasoning loop, allowing agents to write, retrieve, and revise information across rounds. Thus, scalable MAD requires three complementary capabilities: structured trajectory representation, learned utility-aware selection, and dynamic intra-debate memory access.

To address these challenges, we propose \textbf{\name{}} (\underline{\textbf{H}}ypergraph-\underline{\textbf{E}}nhanced \underline{\textbf{D}}ebate with \underline{\textbf{G}}ated \underline{\textbf{E}}vidence memory), a framework that maintains a structured, shared memory throughout an ongoing debate. \name{} represents the debate trajectory as a semantic hypergraph, where hyperedges capture higher-order dependencies among claims and agents beyond pairwise relations. Within each hyperedge, a relational graph preserves fine-grained argumentative structure, including support, refutation, and evidential links. A learned \emph{memory selection module} (MSM), pretrained with supervised learning and refined through reinforcement learning using multi-agent feedback, routes each candidate memory item to \textsc{store}, \textsc{backup}, or \textsc{omit}. Agents retrieve and update the shared memory across rounds, preserving useful dependencies while reducing redundancy and cost. Figure~\ref{fig:teaser} illustrates how \name{} preserves previously established evidence and rejected hypotheses across debate rounds, preventing agents from repeatedly exploring unproductive reasoning paths. By making relevant information available through structured memory, \name{} enables agents to build on earlier conclusions, reduce redundant discussions, and converge on the correct answer in fewer rounds.

Empirically, across seven benchmarks spanning mathematical reasoning, knowledge conflicts, and long-context QA, \name{} consistently outperforms strong baselines. Compared with the best multi-agent system, \name{} improves accuracy by 5.0 points on average, with gains of up to 8.2 points on conflict-heavy tasks such as ConflictQA~\citep{Xie2024KnowledgeConflict}. These gains increase to 11.5 and 16.2 points over the strongest advanced single-agent baseline and the vanilla single-agent using the same backbone, respectively. \name{} also reduces memory usage by 48\% and the number of debate rounds by 34\% relative to the strongest multi-agent baseline, demonstrating its ability to improve reasoning efficiency while maintaining a more compact memory.

\textbf{Overall, our contributions are:} \textbf{(i)}~a shared hypergraph-based memory that captures higher-order semantic dependencies and fine-grained argumentative relations across agents; \textbf{(ii)}~a learned memory selection module, trained with supervised learning and refined through reinforcement learning from debate outcomes, that selectively manages information based on its evolving utility; and \textbf{(iii)}~an intra-debate memory framework that enables agents to retrieve and update relevant information across rounds.

%% file: sections/3_related_work.tex
\section{Related Work}
\label{sec:related_work}

\paragraph{Multi-agent debate and debate memory.}
Multi-agent debate (MAD) improves reasoning by allowing agents to iteratively propose, critique, and refine their answers~\citep{du2024improving,liang-etal-2024-encouraging,chan2024chateval,chen-etal-2024-reconcile,liu2025breaking}. Most MAD frameworks, however, preserve earlier interactions only as an unstructured transcript, making claims, revisions, and cross-round dependencies difficult to track. MeMAD retrieves summaries of completed debates to support new tasks~\citep{ling2025memad}, while G-Memory models cross-trial collaboration through pairwise graph structures~\citep{zhang2025gmemory}. Recent methods more directly address intra-debate memory: MAD-M$^2$ masks potentially erroneous information inherited from previous rounds~\citep{tian2026multiagentdebatememorymasking}, and EquiMem calibrates updates to a shared memory based on agent reliability~\citep{meng2026equimemcalibratingsharedmemory}. Unlike these approaches, \name{} jointly structures the ongoing debate as a shared semantic hypergraph, preserves higher-order dependencies among multiple agents and claims, and exposes this memory for retrieval and revision throughout the debate.

\paragraph{Memory representation and selection.}
Existing language-agent memories include flat stores~\citep{zhong2024memorybank,packer2023memgpt}, retrieval-based architectures~\citep{qian2025memorag,wang-han-2025-proprag}, and learned memory-management methods such as Memory-R1, MemAgent, and CoMAM~\citep{yan2025memory,yu2026memagent,mao2026collaborative}. Flat memories provide limited relational context, while conventional graph memories generally represent only pairwise connections. HyperGraphRAG~\citep{luo2025hypergraphrag} and PRoH~\citep{zai2025proh} capture $N$-ary relations, but focus on retrieving external knowledge rather than maintaining an evolving argumentative state. Moreover, existing learned memory managers primarily target single-agent long-context reasoning or general memory pipelines. In contrast, \name{} combines a semantic hypergraph with fine-grained relational graphs and a learned selection module trained from debate outcomes, enabling utility-aware management of higher-order information within an ongoing multi-agent debate.

%% file: sections/4_memory.tex
\section{\name{}}
\label{sec:hypergraph_memory}

\begin{figure*}
    \centering
    \includegraphics[width=\textwidth]{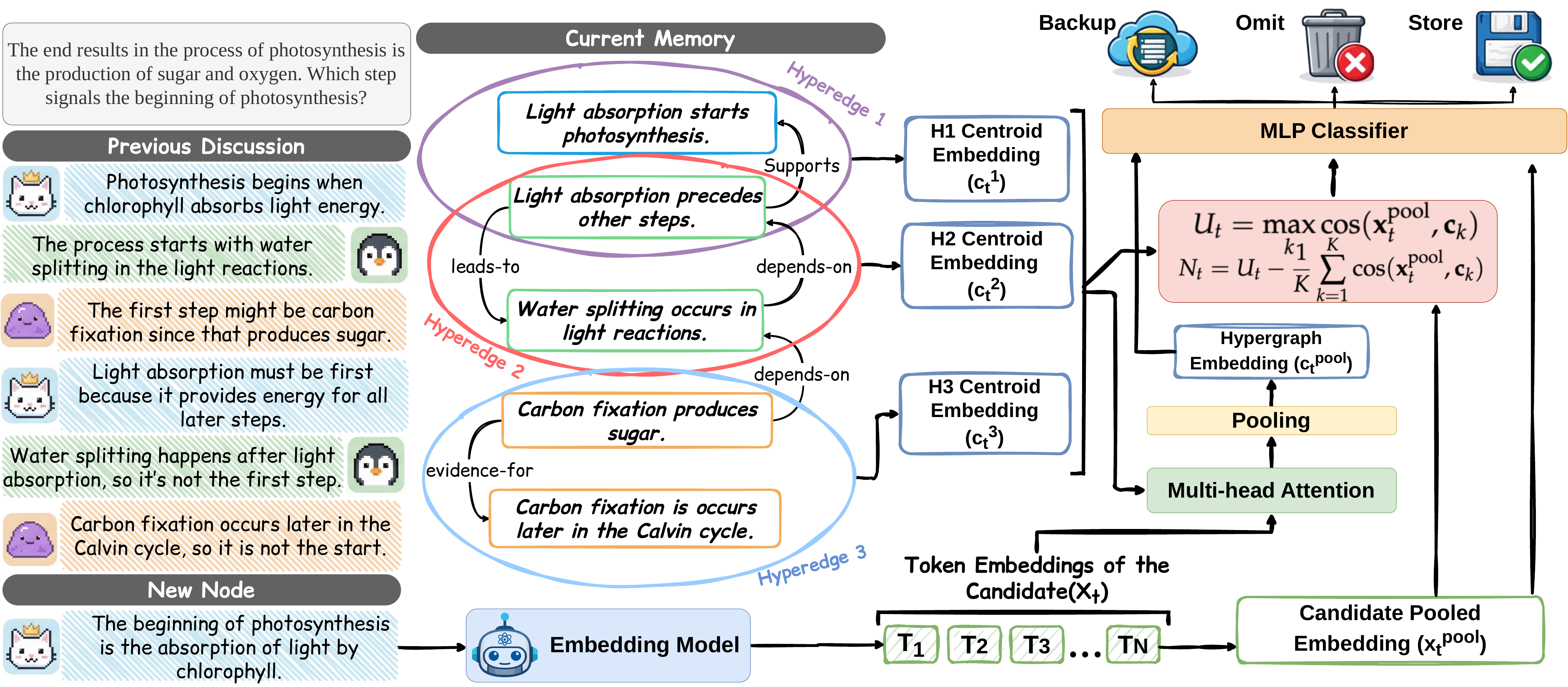}
    \caption{\textbf{Overview of \name{}'s memory update
    pipeline.} The current debate is organized as a
    semantic hypergraph, while the memory selection
    module uses candidate features and cross-attention
    over hyperedge centroids to route new information
    to \textsc{store}, \textsc{backup}, or \textsc{omit}.}
    \label{fig:pipeline}
    \vspace{-1em}
\end{figure*}

As shown in Figure~\ref{fig:pipeline}, \name{} maintains a shared two-level memory throughout an ongoing debate. Agent responses are decomposed into atomic candidate nodes, and a learned memory selection module (MSM) controls their retention. At each candidate step $t$, MSM assigns $v_t$ to \textsc{store}, \textsc{backup}, or \textsc{omit}: \textsc{store} admits the candidate to the structured memory, \textsc{backup} defers the decision for later reconsideration, and \textsc{omit} excludes it from retention. Stored nodes are grouped into semantic hyperedges, while directed relational graphs within each hyperedge preserve fine-grained argumentative links. All agents can retrieve this shared memory during reasoning, and memory maintenance is performed at the end of each debate round.

Formally, the memory at step $t$ consists of a hypergraph $\mathcal{H}^{(t)}=(\mathcal{V}^{(t)},\mathcal{E}^{(t)})$ and directed relational graphs $\{\mathcal{G}^{(t)}_k\}_{k=1}^{|\mathcal{E}^{(t)}|}$. Each node $v\in\mathcal{V}^{(t)}$ represents an atomic debate element, such as a claim, counterclaim, evidence statement, or intermediate conclusion. Each hyperedge $e^{(t)}_k\in\mathcal{E}^{(t)}$ groups semantically related nodes, with an associated graph $\mathcal{G}^{(t)}_k=(\mathcal{N}^{(t)}_k,\mathcal{R}^{(t)}_k)$, where $\mathcal{N}^{(t)}_k=\{v\in\mathcal{V}^{(t)}\mid v\text{ belongs to }e^{(t)}_k\}$ and $\mathcal{R}^{(t)}_k$ contains typed, directed relations between its members. Higher-order structure arises from the joint membership of multiple claims in a hyperedge, while pairwise relations specify how individual claims support, refute, or depend on one another. A node may belong to multiple hyperedges without being duplicated, connecting related topics through shared evidence or conclusions. Thus, connections across hyperedges arise through overlapping membership rather than separate cross-hyperedge relation types.

Because the memory is updated online~\citep{zhang2025gmemory,zhou2025improving}, its nodes, hyperedges, relations, and cardinalities refer to the current debate state. Unlike static retrieval memories~\citep{luo2025hypergraphrag,wang-han-2025-proprag}, it is accessed and revised throughout the current interaction, preserving intermediate reasoning for use in subsequent rounds.

\subsection{Memory Construction and Maintenance}
\label{sec:memory_construction}

For each node $v$ with text $x_v$, we compute an $\ell_2$-normalized embedding and a sparse semantic signature:
\begin{equation}
\mathbf{h}_v=f_{\mathrm{emb}}(x_v), \qquad
\sigma_v=g_{\mathrm{sig}}(x_v),
\end{equation}
where $\sigma_v$ contains salient lexical features such as entities, noun phrases, numerical values, and content lemmas. The embedding captures semantic similarity, while the signature preserves explicit lexical cues that may distinguish closely related claims.

At debate step $t$, each hyperedge $e_k^{(t)}$ has member set $\mathcal{N}_k^{(t)}$. We maintain the unnormalized sum of its member embeddings and compute the mean embedding as:
\begin{equation}
\mathbf{s}_k^{(t)}
=
\sum_{u\in\mathcal{N}_k^{(t)}}\mathbf{h}_u,
\qquad
\bar{\mathbf{h}}_k^{(t)}
=
\frac{\mathbf{s}_k^{(t)}}{|\mathcal{N}_k^{(t)}|}.
\end{equation}
The normalized centroid and union signature are then:
\begin{equation}
\mathbf{c}_k^{(t)}
=
\frac{\mathbf{s}_k^{(t)}}{\|\mathbf{s}_k^{(t)}\|_2},
\qquad
\sigma_k^{(t)}
=
\bigcup_{u\in\mathcal{N}_k^{(t)}}\sigma_u.
\end{equation}
We cache the unnormalized embedding sums to support incremental centroid updates and hyperedge merging without recomputing embeddings for unchanged nodes.

For a node $v$ admitted by MSM, we compare its signature and embedding with each existing hyperedge:
\begin{equation}
J_{vk}^{(t)}=
\frac{|\sigma_v\cap\sigma_k^{(t)}|}
     {|\sigma_v\cup\sigma_k^{(t)}|},
\qquad
S_{vk}^{(t)}=
\mathbf{h}_v^\top\mathbf{c}_k^{(t)}.
\end{equation}
A hyperedge is compatible only when both signature overlap and semantic similarity exceed their respective thresholds:
\begin{equation}
\label{eq:hyperedge-compatibility}
\mathcal{M}_t(v)=
\left\{
e_k^{(t)}\in\mathcal{E}^{(t)}
\;\middle|\;
\begin{aligned}
J_{vk}^{(t)} &\geq \tau_{\mathrm{sig}},\\
S_{vk}^{(t)} &\geq \tau_{\mathrm{emb}}
\end{aligned}
\right\}.
\end{equation}
If $\mathcal{M}_t(v)=\varnothing$, we create a new hyperedge; otherwise, $v$ joins every compatible hyperedge. Compatibility indicates related subject matter rather than agreement: supporting and conflicting claims can therefore belong to the same group, with their argumentative relationship represented separately. Multi-membership allows a claim relevant to several topics to be accessed through each corresponding hyperedge. For every affected hyperedge, we update the cached embedding sum, normalize it to obtain the centroid, and update the union signature.

Within each hyperedge, the directed, typed relational graph $\mathcal{G}_k^{(t)}$ captures fine-grained connections using five relation types: \textsc{supports}, \textsc{refutes}, \textsc{evidence-for}, \textsc{depends-on}, and \textsc{leads-to}, where $(v_i,r,v_j)$ denotes relation $r$ from $v_i$ to $v_j$. When a node joins a hyperedge, an LLM-based module using \texttt{GPT-4o-mini}~\citep{openai2024gpt4technicalreport} compares it with existing members and predicts one of these relations or no relation using Prompt~\ref{prompt:relation_configuration}. Hyperedge membership alone does not imply a relational link. This local update preserves argumentative structure without recomputing relations between unchanged node pairs.

\paragraph{Online Maintenance.}
Candidate-level routing decisions are consolidated at the end of each debate round. We first reconsider \textsc{backup} nodes for promotion to \textsc{store}. Promoted and newly stored nodes are assigned to compatible hyperedges using Eq.~\eqref{eq:hyperedge-compatibility}, or initialize new hyperedges when no match exists. We then merge redundant hyperedges, prune those with persistently low utility, and reassign their useful nodes when possible. Merging combines overlapping memberships without duplicating shared nodes. To limit overhead, node embeddings and signatures are cached, previously inferred relations between unchanged node pairs are retained, and only affected hyperedge representations and relational graphs are updated. Appendix~\ref{app:online-maintenance} provides the complete procedure.






\subsection{Two-Level Memory Retrieval}
\label{sec:memory_read}

Retrieval follows the two-level memory structure: it first identifies relevant semantic groups and then selects individual claims together with their argumentative relations. At debate step $t$, the agent forms a query $q$, encoded as an embedding $\mathbf{h}_q$ and a sparse signature $\sigma_q$ using the same functions as memory nodes.

At the hyperedge level, we select up to $K$ hyperedges by centroid similarity, subject to signature compatibility:
\begin{equation}
\mathcal{E}_q^{(t)}
=
\operatorname{TopK}_{\substack{
e_k^{(t)}\in\mathcal{E}^{(t)}\\
J(\sigma_q,\sigma_k^{(t)})\geq\tau_{\mathrm{sig}}
}}
\cos\!\left(\mathbf{h}_q,\mathbf{c}_k^{(t)}\right).
\end{equation}
The signature gate requires overlap in salient lexical cues, while centroid similarity ranks the compatible groups by semantic relevance. This coarse selection focuses subsequent retrieval on relevant portions of the debate memory.

Within each selected hyperedge, we retrieve up to $M$ member nodes by embedding similarity:
\begin{equation}
\mathcal{V}_q^{(k,t)}
=
\operatorname{TopM}_{v\in\mathcal{N}_k^{(t)}}
\cos\!\left(\mathbf{h}_q,\mathbf{h}_v\right).
\end{equation}
This second stage selects the most relevant claims within each group rather than returning every member. Because nodes may belong to multiple hyperedges, shared evidence or intermediate conclusions can be selected through different semantic groups.

For each $e_k^{(t)}\in\mathcal{E}_q^{(t)}$, we return the induced subgraph
$\mathcal{G}_k^{(t)}[\mathcal{V}_q^{(k,t)}]$,
containing the selected nodes and all existing directed, typed relations whose endpoints are both selected. Thus, retrieval preserves how the returned claims support, refute, or depend on one another, rather than exposing them only as isolated text fragments. Relations involving unselected nodes are excluded.

The resulting subgraphs form a memory readout containing relevant claims, evidence, refutations, and dependencies. The retrieval budget permits at most $KM$ node selections before accounting for overlapping membership, bounding the amount of memory returned while allowing evidence to be drawn from multiple related groups.

%% file: sections/5_msm.tex
\subsection{Memory Selection Module}
\label{sec:msm}

Given a candidate node $v_t$, extracted as an atomic fact from an agent response using \texttt{GPT-4o-mini}~\citep{openai2024gpt4technicalreport} and Prompt~\ref{prompt:atomic_fact_extraction}, MSM predicts
$a_t \in \{\textsc{store},\textsc{backup},\textsc{omit}\}$
based on the candidate and current memory. Let
$\mathbf{X}_t \in \mathbb{R}^{n_t \times d_e}$ denote its token representations and
$\mathbf{x}_t=\operatorname{Pool}(\mathbf{X}_t)$ its pooled representation.

For each hyperedge $e_k^{(t)}$, we first compute the candidate-to-hyperedge alignment:
\begin{equation}
u_{t,k}
=
\cos\!\left(\mathbf{x}_t,\mathbf{c}_k^{(t)}\right).
\end{equation}
We then construct the alignment and specificity features:
\begin{equation}
\boldsymbol{\phi}_t^{\mathrm{UN}}=[U_t,N_t],
\end{equation}
where
\begin{equation}
\label{eq:UN}
\begin{aligned}
U_t
&=
\max_k u_{t,k},\\
N_t
&=
U_t-
\frac{1}{|\mathcal{E}^{(t)}|}
\sum_{k=1}^{|\mathcal{E}^{(t)}|}u_{t,k}.
\end{aligned}
\end{equation}
Here, $u_{t,k}$ measures the alignment of candidate $v_t$ with hyperedge $e_k^{(t)}$, $U_t$ measures its strongest hyperedge alignment, and $N_t$ measures the specificity of that alignment relative to other hyperedges. The hyperedge-specific scores $u_{t,k}$ are also used for utility-based pruning, as detailed in Appendix~\ref{app:online-maintenance}.

To incorporate memory context, we stack the current hyperedge centroids into
$\mathbf{C}_t\in\mathbb{R}^{K_t\times d_e}$,
where $K_t=|\mathcal{E}^{(t)}|$.
We project the candidate and centroid representations into a shared hidden dimension $d_h$:
\begin{equation}
\widetilde{\mathbf{X}}_t
=
\mathbf{X}_t\mathbf{W}_x,
\qquad
\widetilde{\mathbf{C}}_t
=
\mathbf{C}_t\mathbf{W}_c.
\end{equation}
Candidate-to-memory cross-attention is then computed as:
\begin{equation}
\mathbf{Z}_t
=
\operatorname{MultiHeadAttn}
\left(
\widetilde{\mathbf{X}}_t,
\widetilde{\mathbf{C}}_t,
\widetilde{\mathbf{C}}_t
\right).
\end{equation}
The projected candidate and context representations are obtained by pooling:
\begin{equation}
\mathbf{x}_t^{\mathrm{pool}}
=
\operatorname{Pool}(\widetilde{\mathbf{X}}_t),
\qquad
\mathbf{c}_t^{\mathrm{pool}}
=
\operatorname{Pool}(\mathbf{Z}_t).
\end{equation}

The routing representation combines utility features, candidate semantics, and hypergraph context:
\begin{equation}
\boldsymbol{\phi}_t
=
\left[
\boldsymbol{\phi}_t^{\mathrm{UN}}
\;\middle\|\;
\mathbf{x}_t^{\mathrm{pool}}
\;\middle\|\;
\mathbf{c}_t^{\mathrm{pool}}
\right].
\end{equation}
The action distribution is:
\begin{equation}
\pi_\theta(a_t\mid\boldsymbol{\phi}_t)
=
\operatorname{Softmax}
\left(
\mathbf{W}_\pi
\operatorname{MLP}(\boldsymbol{\phi}_t)
+\mathbf{b}_\pi
\right).
\end{equation}
Here, $\operatorname{MLP}$ comprises two hidden linear layers, while $\mathbf{W}_\pi$ and $\mathbf{b}_\pi$ parameterize the third, output layer. Further architectural details are provided in Appendix~\ref{app:classifier_architecture}.

\subsection{Training the Memory Selection Module}
\label{sec:msm_training}

MSM is a lightweight three-layer MLP with normalization and dropout that predicts
$\{\textsc{store},\textsc{backup},\textsc{omit}\}$ from the candidate representation and current hypergraph context. Architectural details are provided in Appendix~\ref{app:classifier_architecture}.

\paragraph{Data and setup.}
We construct training data from simulated multi-agent debates on the HotpotQA FullWiki training split~\citep{yang2018hotpotqa}. For each sample, we provide the HotpotQA question and context to GPT-4o~\citep{openai2024gpt4technicalreport}, which generates a task-specific roster of 3--6 agents with complementary roles, goals, model sizes, and turn order using Prompt~\ref{prompt:roster_generation}. We then simulate the full debate and log all agent outputs, retrieved evidence, memory references, routing decisions, and final answers. For Stage~I, these trajectories are analyzed post hoc to determine whether each candidate memory item is later reused, supports conflict resolution, or contributes to the final answer, yielding 6.5K pseudo-labeled routing examples. For Stage~II, we use an additional 1K HotpotQA training questions to simulate online debates and refine the policy through interaction. Appendix~\ref{sec:msm_data_generation} provides full data-generation details.

\paragraph{Stage I: Supervised pretraining.}
We derive labels post hoc based on whether an item is reused, resolves a conflict, contributes to the final output, or is redundant. Frequently useful and non-redundant items are labeled \textsc{store}, potentially useful items \textsc{backup}, and unused items \textsc{omit}. MSM is trained with categorical cross-entropy and a class-balancing loss that prevents collapse toward the dominant routing action. This stage provides a general initialization for utility-aware routing based on relevance, specificity, and redundancy.

\paragraph{Stage II: Reinforcement learning.}
We further optimize MSM with PPO~\citep{schulman2017proximalpolicyoptimizationalgorithms} in an online multi-agent debate environment. PPO is suitable because routing decisions have delayed trajectory-level effects: a stored item may become useful several rounds later, while excessive retention gradually increases memory cost. Its value estimates support delayed credit assignment, and clipped updates preserve stability around the supervised initialization. PPO also handles MSM's discrete actions and directly optimizes non-differentiable objectives, including task accuracy, future memory use, and memory size. The reward combines task success, routing utility based on later item usage, and a penalty for memory growth. Training uses the clipped PPO objective with value loss and entropy regularization to support long-horizon adaptation without collapsing to a single action.

Full details on pseudo-label construction, reward design, and optimization are provided in Appendix~\ref{app:msm_train_proc}.

%% file: sections/6_experiments.tex
\section{Experiments and Results}

\subsection{Experimental Settings}

\begin{table*}[t]
    \centering
    \resizebox{\textwidth}{!}{%
    \begin{tabular}{l|ccccccc}
        \hline
        \textbf{Method} 
        & \textbf{MATH500} 
        & \textbf{GPQA} 
        & \textbf{FaithEval} 
        & \textbf{ConflictQA} 
        & \textbf{NovelQA} 
        & \textbf{HotpotQA} 
        & \textbf{MMLU-Pro} \\
        \hline
        
        \rowcolor{gray!20}
        \multicolumn{8}{c}{\textbf{Single-Agent (Vanilla)}} \\ \hline
        
        GPT-5.4
        & \score{57.1}{0.2}
        & \score{56.2}{0.4}
        & \score{72.7}{0.3}
        & \score{48.9}{0.5}
        & \score{62.4}{0.4}
        & \score{45.3}{0.3}
        & \score{85.2}{0.2} \\
        
        GPT-4o
        & \score{53.2}{0.3}
        & \score{44.2}{0.5}
        & \score{66.4}{0.4}
        & \score{41.5}{0.6}
        & \score{58.1}{0.3}
        & \score{38.9}{0.4}
        & \score{77.6}{0.3} \\
        
        GPT-4o-mini
        & \score{49.8}{0.4}
        & \score{33.8}{0.6}
        & \score{61.4}{0.5}
        & \score{29.2}{0.7}
        & \score{53.3}{0.4}
        & \score{29.4}{0.5}
        & \score{66.8}{0.4} \\
        
        Qwen3.5-27B
        & \score{51.4}{0.3}
        & \score{48.9}{0.4}
        & \score{69.2}{0.3}
        & \score{36.8}{0.6}
        & \score{57.6}{0.5}
        & \score{39.8}{0.4}
        & \score{83.4}{0.2} \\
        
        \hline
        
        \rowcolor{gray!20}
        \multicolumn{8}{c}{
            \textbf{Single-Agent (Advanced)} (\texttt{GPT-4o-mini})
        } \\ \hline
        
        Self-Refine
        & \score{53.7}{0.5}
        & \score{37.6}{0.6}
        & \score{64.9}{0.4}
        & \score{33.4}{0.7}
        & \score{54.1}{0.5}
        & \score{30.7}{0.6}
        & \score{71.2}{0.3} \\
        
        Self-Consistency
        & \score{54.8}{0.4}
        & \score{36.5}{0.7}
        & \score{65.7}{0.5}
        & \score{34.1}{0.6}
        & \score{53.8}{0.6}
        & \score{31.2}{0.4}
        & \score{72.6}{0.4} \\
        
        Reflexion
        & \score{55.6}{0.5}
        & \score{38.2}{0.5}
        & \score{66.5}{0.6}
        & \score{35.3}{0.8}
        & \score{54.8}{0.4}
        & \score{31.9}{0.5}
        & \score{73.8}{0.3} \\

        \hline

        \rowcolor{gray!20}
        \multicolumn{8}{c}{
            \textbf{Multi-Agent} (\texttt{GPT-4o-mini})
        } \\ \hline
        
        MAD
        & \score{55.4}{0.6}
        & \score{38.8}{0.5}
        & \score{67.4}{0.6}
        & \score{36.7}{0.7}
        & \score{55.5}{0.5}
        & \score{32.4}{0.4}
        & \score{74.6}{0.4} \\
        
        MAD-M$^2$
        & \score{55.9}{0.4}
        & \score{39.4}{0.6}
        & \score{67.8}{0.5}
        & \score{37.6}{0.8}
        & \score{56.5}{0.6}
        & \score{32.8}{0.5}
        & \score{75.4}{0.3} \\
        
        MoA
        & \score{56.2}{0.5}
        & \score{39.7}{0.4}
        & \score{68.3}{0.6}
        & \score{38.1}{0.7}
        & \score{57.2}{0.5}
        & \score{33.1}{0.4}
        & \score{75.8}{0.3} \\
        
        MeMAD
        & \score{57.8}{0.6}
        & \score{41.5}{0.5}
        & \score{70.2}{0.4}
        & \score{40.6}{0.9}
        & \score{58.7}{0.6}
        & \score{34.9}{0.5}
        & \score{77.1}{0.4} \\
        
        G-Memory
        & \score{58.6}{0.5}
        & \score{42.8}{0.6}
        & \score{73.6}{0.5}
        & \score{46.3}{0.8}
        & \score{58.9}{0.7}
        & \score{36.1}{0.4}
        & \score{78.4}{0.3} \\
        
        G-Memory + EquiMem
        & \score{59.4}{0.4}
        & \score{43.7}{0.5}
        & \score{74.9}{0.4}
        & \score{47.8}{0.7}
        & \score{59.8}{0.5}
        & \score{37.0}{0.6}
        & \score{79.0}{0.4} \\
        
        \rowcolor{cyan!12}
        \textbf{\name{}}
        & \bestscore{62.2}{0.2}
        & \bestscore{47.7}{0.3}
        & \bestscore{79.8}{0.3}
        & \bestscore{56.0}{0.4}
        & \bestscore{66.0}{0.2}
        & \bestscore{42.4}{0.2}
        & \bestscore{82.7}{0.2} \\

        \hline
        
        \rowcolor{gray!20}
        \multicolumn{8}{c}{
            \textbf{Single-Agent (Advanced)} (\texttt{Qwen3.5-27B})
        } \\ \hline
        
        Self-Refine
        & \score{54.2}{0.4}
        & \score{51.3}{0.5}
        & \score{72.1}{0.4}
        & \score{39.6}{0.6}
        & \score{59.8}{0.5}
        & \score{42.1}{0.4}
        & \score{86.0}{0.2} \\
        
        Self-Consistency
        & \score{55.1}{0.3}
        & \score{52.0}{0.6}
        & \score{72.8}{0.5}
        & \score{40.3}{0.7}
        & \score{60.2}{0.4}
        & \score{42.6}{0.5}
        & \score{86.9}{0.3} \\
        
        Reflexion
        & \score{56.0}{0.4}
        & \score{53.1}{0.5}
        & \score{73.6}{0.4}
        & \score{41.5}{0.6}
        & \score{61.0}{0.5}
        & \score{43.4}{0.4}
        & \score{88.1}{0.2} \\
        
        \hline

        \rowcolor{gray!20}
        \multicolumn{8}{c}{
            \textbf{Multi-Agent} (\texttt{Qwen3.5-27B})
        } \\ \hline
        
        MAD
        & \score{57.3}{0.4}
        & \score{52.4}{0.5}
        & \score{73.0}{0.4}
        & \score{42.1}{0.6}
        & \score{59.1}{0.5}
        & \score{42.0}{0.4}
        & \score{85.8}{0.3} \\
        
        MAD-M$^2$
        & \score{57.7}{0.5}
        & \score{53.2}{0.4}
        & \score{73.8}{0.5}
        & \score{43.1}{0.7}
        & \score{60.0}{0.4}
        & \score{42.5}{0.5}
        & \score{86.3}{0.2} \\
        
        MoA
        & \score{58.1}{0.4}
        & \score{53.6}{0.5}
        & \score{74.2}{0.3}
        & \score{43.7}{0.6}
        & \score{60.4}{0.5}
        & \score{42.8}{0.4}
        & \score{86.6}{0.3} \\
        
        MeMAD
        & \score{59.3}{0.5}
        & \score{56.8}{0.4}
        & \score{76.1}{0.5}
        & \score{47.0}{0.7}
        & \score{61.7}{0.4}
        & \score{44.2}{0.5}
        & \score{87.4}{0.2} \\
        
        G-Memory
        & \score{60.2}{0.4}
        & \score{57.6}{0.5}
        & \score{78.3}{0.4}
        & \score{50.1}{0.6}
        & \score{62.0}{0.5}
        & \score{44.6}{0.4}
        & \score{88.2}{0.3} \\
        
        G-Memory + EquiMem
        & \score{61.0}{0.5}
        & \score{58.7}{0.4}
        & \score{79.5}{0.3}
        & \score{51.8}{0.7}
        & \score{63.1}{0.4}
        & \score{45.4}{0.5}
        & \score{88.7}{0.2} \\
        
        \rowcolor{cyan!12}
        \textbf{\name{}}
        & \bestscore{64.3}{0.2}
        & \bestscore{62.2}{0.3}
        & \bestscore{83.6}{0.3}
        & \bestscore{57.9}{0.3}
        & \bestscore{67.6}{0.4}
        & \bestscore{49.2}{0.1}
        & \bestscore{90.3}{0.2} \\
        
        \hline
    \end{tabular}
    }
    \caption{
        Accuracy (\%) across seven benchmarks. Each value reports the mean over \textbf{three runs}, with red subscripts indicating the corresponding error. All multi-agent methods use a three-agent topology. \name{} performs best under both backbone models, with particularly strong gains on FaithEval, ConflictQA, and NovelQA.
    }
    \label{tab:main_results}
    \vspace{-1em}
\end{table*}

\begin{table*}[t]
    \centering
    \resizebox{\textwidth}{!}{%
    \begin{tabular}{l|ccccccc}
        \hline
        \textbf{Memory} 
        & \textbf{MATH500} 
        & \textbf{GPQA} 
        & \textbf{FaithEval} 
        & \textbf{ConflictQA} 
        & \textbf{NovelQA} 
        & \textbf{HotpotQA} 
        & \textbf{MMLU-Pro} \\
        \hline
        
        \rowcolor{gray!20}
        \multicolumn{8}{c}{\textbf{AutoGen~\citeyear{wu2024autogen}}} \\ 
        \hline
        
        No-memory
        & \score{55.5}{0.5}
        & \score{38.9}{0.6}
        & \score{68.0}{0.4}
        & \score{36.5}{0.7}
        & \score{56.3}{0.5}
        & \score{32.7}{0.4}
        & \score{74.7}{0.3} \\
        
        Voyager
        & \score{57.3}{0.4}
        & \score{40.7}{0.5}
        & \score{69.8}{0.6}
        & \score{38.6}{0.8}
        & \score{58.1}{0.4}
        & \score{34.2}{0.5}
        & \score{76.3}{0.3} \\
        
        Memory Bank
        & \score{54.9}{0.6}
        & \score{38.2}{0.7}
        & \score{67.3}{0.5}
        & \score{37.2}{0.6}
        & \score{57.1}{0.6}
        & \score{33.5}{0.4}
        & \score{75.3}{0.4} \\
        
        Generative
        & \score{57.8}{0.5}
        & \score{41.4}{0.6}
        & \score{70.5}{0.4}
        & \score{39.3}{0.7}
        & \score{58.9}{0.5}
        & \score{34.7}{0.6}
        & \score{77.1}{0.3} \\
        
        \midrule
        
        MetaGPT-M
        & \score{58.4}{0.4}
        & \score{42.1}{0.5}
        & \score{71.1}{0.5}
        & \score{40.2}{0.6}
        & \score{59.6}{0.4}
        & \score{35.3}{0.5}
        & \score{77.9}{0.3} \\
        
        MacNet-M
        & \score{56.5}{0.5}
        & \score{40.3}{0.6}
        & \score{69.1}{0.4}
        & \score{38.3}{0.8}
        & \score{57.9}{0.6}
        & \score{34.1}{0.5}
        & \score{76.1}{0.4} \\
        
        ChatDev-M
        & \score{54.1}{0.6}
        & \score{37.5}{0.7}
        & \score{66.4}{0.5}
        & \score{35.6}{0.7}
        & \score{56.0}{0.5}
        & \score{32.6}{0.6}
        & \score{74.2}{0.4} \\
        
        \midrule

        MAD-M$^2$
        & \score{58.8}{0.5}
        & \score{42.7}{0.4}
        & \score{71.6}{0.5}
        & \score{41.1}{0.7}
        & \score{59.8}{0.4}
        & \score{35.5}{0.5}
        & \score{78.2}{0.3} \\
        
        MeMAD
        & \score{59.2}{0.4}
        & \score{43.1}{0.5}
        & \score{72.1}{0.4}
        & \score{41.9}{0.6}
        & \score{60.0}{0.5}
        & \score{35.6}{0.4}
        & \score{78.6}{0.3} \\
        
        G-Memory
        & \score{60.3}{0.4}
        & \score{44.6}{0.5}
        & \score{75.9}{0.4}
        & \score{48.0}{0.6}
        & \score{60.5}{0.5}
        & \score{36.1}{0.4}
        & \score{80.1}{0.3} \\
        
        G-Memory + EquiMem
        & \score{60.9}{0.3}
        & \score{45.4}{0.4}
        & \score{77.0}{0.4}
        & \score{49.6}{0.5}
        & \score{61.4}{0.4}
        & \score{37.7}{0.3}
        & \score{80.8}{0.3} \\
        
        \rowcolor{cyan!12}
        \textbf{\name{}}
        & \bestscore{62.9}{0.2}
        & \bestscore{48.3}{0.3}
        & \bestscore{80.7}{0.4}
        & \bestscore{57.8}{0.3}
        & \bestscore{67.2}{0.3}
        & \bestscore{42.3}{0.3}
        & \bestscore{83.5}{0.1} \\
        
        \hline

        \rowcolor{gray!20}
        \multicolumn{8}{c}{\textbf{DyLAN~\citeyear{liu2024a}}} \\ 
        \hline
        
        No-memory
        & \score{53.2}{0.6}
        & \score{36.8}{0.7}
        & \score{65.7}{0.5}
        & \score{34.6}{0.8}
        & \score{54.3}{0.6}
        & \score{31.3}{0.5}
        & \score{72.6}{0.4} \\
        
        Voyager
        & \score{54.9}{0.5}
        & \score{38.4}{0.6}
        & \score{67.4}{0.5}
        & \score{36.5}{0.7}
        & \score{56.0}{0.5}
        & \score{32.6}{0.6}
        & \score{74.1}{0.3} \\
        
        Memory Bank
        & \score{52.6}{0.7}
        & \score{36.2}{0.6}
        & \score{65.0}{0.6}
        & \score{35.1}{0.8}
        & \score{55.1}{0.5}
        & \score{32.1}{0.5}
        & \score{73.0}{0.4} \\
        
        Generative
        & \score{55.3}{0.5}
        & \score{39.0}{0.7}
        & \score{68.0}{0.4}
        & \score{36.9}{0.6}
        & \score{56.6}{0.6}
        & \score{33.0}{0.5}
        & \score{74.6}{0.4} \\
        
        \midrule
        
        MetaGPT-M
        & \score{55.8}{0.4}
        & \score{39.6}{0.6}
        & \score{68.6}{0.5}
        & \score{37.6}{0.7}
        & \score{57.2}{0.5}
        & \score{33.5}{0.4}
        & \score{75.2}{0.3} \\
        
        MacNet-M
        & \score{54.1}{0.6}
        & \score{37.9}{0.5}
        & \score{66.7}{0.6}
        & \score{35.9}{0.8}
        & \score{55.6}{0.5}
        & \score{32.5}{0.6}
        & \score{73.9}{0.4} \\
        
        ChatDev-M
        & \score{51.8}{0.7}
        & \score{35.4}{0.6}
        & \score{64.1}{0.5}
        & \score{33.3}{0.7}
        & \score{53.9}{0.6}
        & \score{30.8}{0.5}
        & \score{71.8}{0.4} \\
        
        \midrule

        MAD-M$^2$
        & \score{56.2}{0.5}
        & \score{40.0}{0.5}
        & \score{68.9}{0.4}
        & \score{38.4}{0.6}
        & \score{57.5}{0.4}
        & \score{33.8}{0.5}
        & \score{75.6}{0.3} \\
        
        MeMAD
        & \score{56.6}{0.5}
        & \score{40.3}{0.6}
        & \score{69.1}{0.4}
        & \score{39.2}{0.7}
        & \score{57.8}{0.5}
        & \score{34.0}{0.4}
        & \score{76.0}{0.3} \\
        
        G-Memory
        & \score{57.5}{0.4}
        & \score{41.6}{0.5}
        & \score{72.4}{0.5}
        & \score{44.8}{0.6}
        & \score{58.2}{0.4}
        & \score{34.4}{0.5}
        & \score{77.3}{0.3} \\
        
        G-Memory + EquiMem
        & \score{58.1}{0.4}
        & \score{42.4}{0.5}
        & \score{73.5}{0.4}
        & \score{46.0}{0.6}
        & \score{59.1}{0.5}
        & \score{35.2}{0.4}
        & \score{78.0}{0.3} \\
        
        \rowcolor{cyan!12}
        \textbf{\name{}}
        & \bestscore{60.6}{0.2}
        & \bestscore{44.8}{0.3}
        & \bestscore{75.1}{0.3}
        & \bestscore{50.7}{0.3}
        & \bestscore{62.4}{0.4}
        & \bestscore{37.4}{0.2}
        & \bestscore{80.1}{0.2} \\
        
        \hline
    \end{tabular}
    }
    \caption{
        Comparison of memory mechanisms under AutoGen and DyLAN. Values are mean accuracy over three runs, with red subscripts indicating standard deviation. \name{} achieves the best performance across all benchmarks. See Appendix~\ref{app:mas_additional} for the results on MacNet and G-Designer.
    }
    \vspace{-1em}
    \label{tab:mas_memory}
\end{table*}

\paragraph{Baselines.}
We compare against vanilla single-agent models (GPT-5.4~\citep{openai2026gpt54}, GPT-4o, GPT-4o-mini~\citep{openai2024gpt4technicalreport}, and Qwen3.5-27B~\citep{qwen35}), self-improvement methods (Self-Refine~\citep{selfrefine}, Self-Consistency~\citep{wang2023selfconsistency}, and Reflexion~\citep{shinn2023reflexion}), and multi-agent baselines (MAD~\citep{liang-etal-2024-encouraging} and MoA~\citep{wang2024moa}). Memory-aware baselines include MeMAD~\citep{ling2025memad}, G-Memory~\citep{zhang2025gmemory}, G-Memory~+~EquiMem~\citep{meng2026equimemcalibratingsharedmemory}, and the perplexity-based variant of MAD-M$^2$~\citep{tian2026multiagentdebatememorymasking}. To test portability, we integrate \name{} into AutoGen~\citep{wu2024autogen}, G-Designer~\citep{zhang2025gdesigner}, DyLAN~\citep{liu2024a}, and MacNet~\citep{qian2025scaling}, and compare against Voyager~\citep{wang2023voyager}, Memory Bank~\citep{zhong2024memorybank}, Generative Agents~\citep{park2023generative}, and native memory variants. See Appendix~\ref{app:baseline_details} for baseline details and Appendix~\ref{app:mas_frameworks} for descriptions of the multi-agent frameworks.

\paragraph{Benchmarks.}
We evaluate on GPQA~\citep{rein2024gpqa} and MATH500~\citep{hendrycks2measuring} for mathematical and scientific reasoning; FaithEval~\citep{ming2024faitheval} and ConflictQA~\citep{Xie2024KnowledgeConflict} for faithfulness under conflicting evidence; NovelQA~\citep{wang2024novelqa} and HotpotQA fullwiki~\citep{yang2018hotpotqa} for long-context and multi-hop reasoning; and MMLU-Pro~\citep{wang2024mmlu} for broad knowledge reasoning. MSM is trained only on the HotpotQA fullwiki training split and evaluated on a disjoint HotpotQA set and six external benchmarks, testing both in-domain robustness and cross-benchmark transfer. We report mean accuracy over three runs. See Appendix~\ref{app:benchmark_details} for benchmark and baseline details, and Appendix~\ref{app:evaluation_prompts} for evaluation prompts.

\paragraph{Experimental setup.}
We use a temperature of $0.7$ and Qwen3-Embedding-4B~\citep{qwen3embedding} for normalized node, query, and hyperedge embeddings. Stage I uses AdamW with learning rate $10^{-4}$, batch size $64$, weight decay $0.01$, and dropout $0.1$; Stage II uses learning rate $10^{-5}$, $\gamma=0.99$, GAE parameter $0.95$, PPO clipping threshold $0.2$, value coefficient $0.5$, entropy coefficient $0.01$, and four epochs per batch. We set $\tau_{\mathrm{sig}}=0.30$ and $\tau_{\mathrm{emb}}=0.70$ for node assignment, $\tau_{\mathrm{merge}}^{\mathrm{sig}}=0.50$ and $\tau_{\mathrm{merge}}^{\mathrm{emb}}=0.85$ for merging, and prune hyperedges with utility below $0.10$ for three rounds. Backup nodes are retained for at most two rounds. During retrieval, we set $K=3$, returning the top three hyperedges with up to top eight nodes per hyperedge ($M=8$). Memory is updated after each debate round. Further details are provided in Appendix~\ref{app:training_details}.


\subsection{Performance Across Reasoning Tasks}
\label{sec:reasoning_performance}

\paragraph{Performance across backbones.}
Table~\ref{tab:main_results} reports results in a three-agent setting with two proposers and one critic. \name{} achieves the best performance among methods using the same backbone on all seven benchmarks. With GPT-4o-mini, it outperforms G-Memory~+~EquiMem by \textbf{5.0 points on average}, with gains ranging from \textbf{2.8} on MATH500 to \textbf{8.2} on ConflictQA. With Qwen3.5-27B, the average gain is \textbf{3.8 points}, ranging from \textbf{1.6} on MMLU-Pro to \textbf{6.1} on ConflictQA. Improvements are particularly strong on \textbf{ConflictQA, NovelQA, and FaithEval}, where agents must retain competing claims, track evidence provenance, and revise conclusions as new information emerges. This pattern suggests that \name{} is most beneficial when reasoning depends on integrating information distributed across agents and debate rounds. The gains of \textbf{5.4} and \textbf{3.8} points on HotpotQA further indicate that intra-hyperedge relations help preserve multi-step evidence chains, whereas the smaller gain on MMLU-Pro suggests less benefit when answers rely primarily on parametric knowledge. Notably, \name{} with Qwen3.5-27B also \textbf{surpasses vanilla GPT-5.4 on all seven benchmarks}, showing that structured debate memory can partly compensate for differences in backbone scale. Appendix~\ref{sec:case_study} provides a qualitative example.

\paragraph{Comparison of memory mechanisms. }
Table~\ref{tab:mas_memory} compares memory mechanisms under AutoGen~\citep{wu2024autogen} and DyLAN~\citep{liu2024a}. \name{} achieves the best result on every benchmark under both frameworks. MAD-M$^2$ consistently improves over native and lightweight memory variants, but remains slightly below MeMAD and substantially behind graph-based memories, suggesting that filtering erroneous content alone cannot fully capture cross-agent dependencies. Under AutoGen, \name{} outperforms the strongest baseline, G-Memory~+~EquiMem, by \textbf{4.3 points on average}, including gains of \textbf{2.0} on MATH500, \textbf{2.9} on GPQA, \textbf{3.7} on FaithEval, and \textbf{8.2} on ConflictQA. Under DyLAN, the average gain is \textbf{2.7 points}, including improvements of \textbf{2.5} on MATH500, \textbf{2.4} on GPQA, and \textbf{4.7} on ConflictQA. The largest gains again occur on \textbf{ConflictQA and NovelQA}, indicating that selective hypergraph memory is particularly effective for preserving and reconciling competing information. The larger gains under AutoGen may reflect its richer interaction pattern, while DyLAN's dynamic orchestration may already reduce some redundant exchanges. Consistent gains across both frameworks show that \name{} is orchestration-agnostic. Appendix~\ref{app:mas_additional} reports results for G-Designer~\citep{zhang2025gdesigner} and MacNet~\citep{qian2025scaling}.

\subsection{Efficiency Analysis}
\label{sec:efficiency}

We evaluate efficiency from two complementary perspectives: \textbf{(i)} total token usage and reasoning rounds, and \textbf{(ii)} persistent memory footprint. Figure~\ref{fig:efficiency} presents representative results, with detailed values for all benchmarks reported in Appendix~\ref{app:detailed_results_efficiency}.

\begin{figure*}[t]
    \centering
    \includegraphics[width=\textwidth]{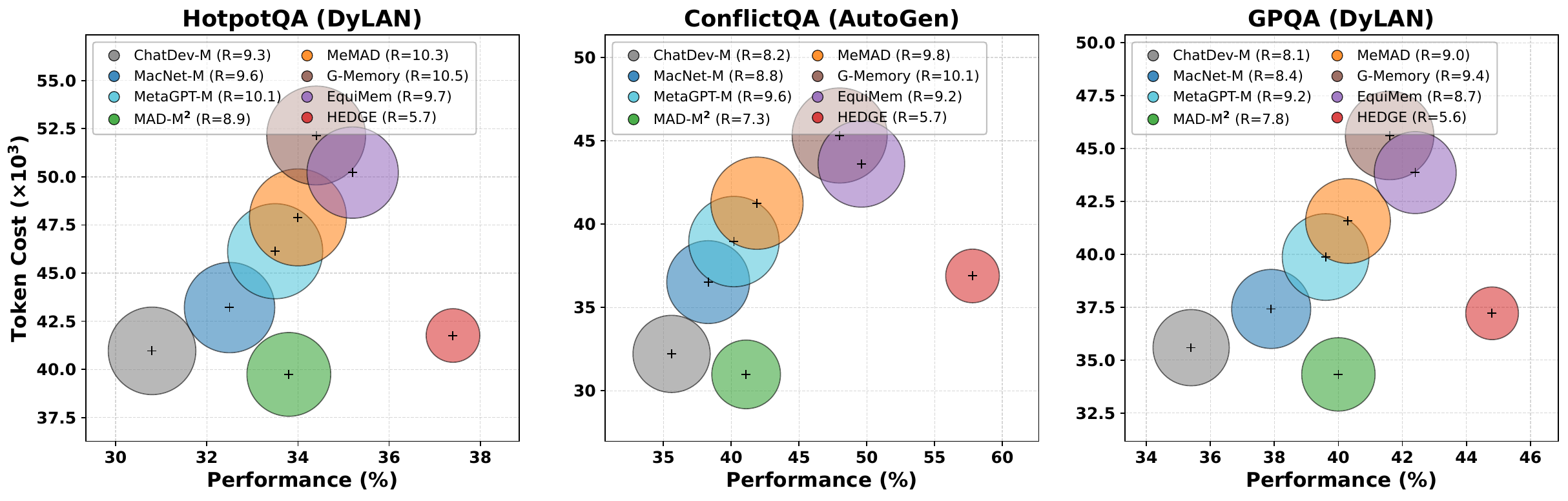}

    \includegraphics[width=\textwidth]{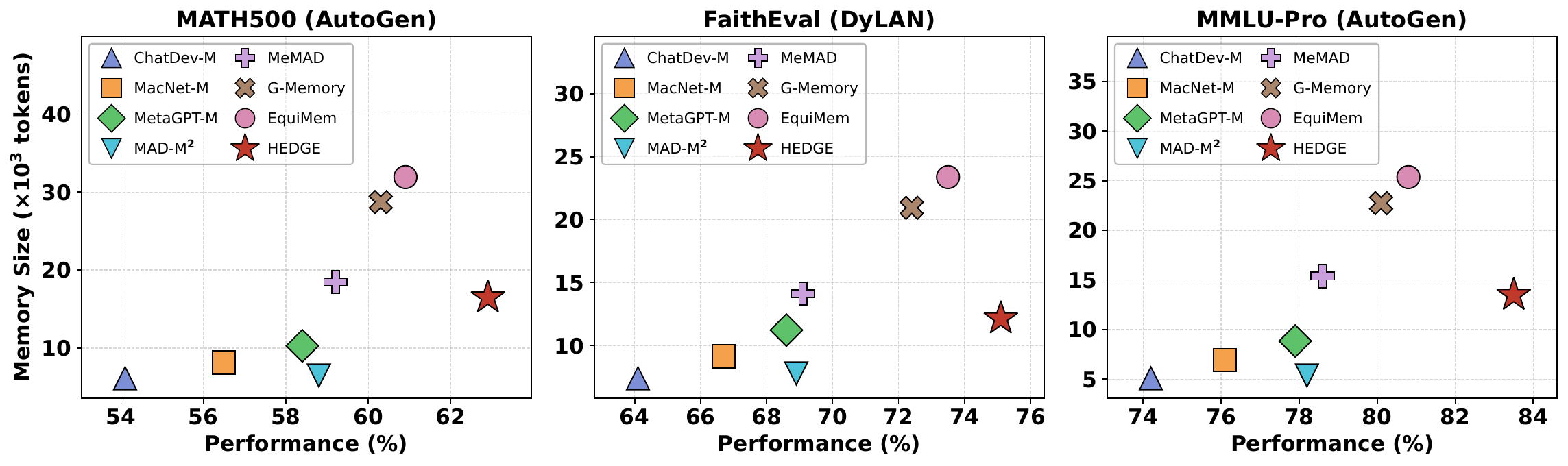}

    \caption{Performance--efficiency trade-offs across representative benchmarks and orchestration frameworks. \textbf{Top:} performance versus total token usage, with bubble size indicating the average number of reasoning rounds. \textbf{Bottom:} performance versus persistent memory footprint. See Appendix~\ref{app:detailed_results_efficiency} for full results across benchmarks.}
    \label{fig:efficiency}
    \vspace{-1em}
\end{figure*}

\paragraph{Token usage and reasoning rounds.}
The top panel of Figure~\ref{fig:efficiency} shows that \name{} achieves a stronger performance--cost trade-off than structured-memory baselines. Compared with G-Memory~+~EquiMem, \name{} reduces token usage by \textbf{15--17\%} and reasoning rounds by \textbf{36--41\%} across the displayed settings, while improving accuracy by \textbf{2.2--8.2 points}. For example, on ConflictQA with AutoGen, \name{} reaches 57.8 accuracy using 36.9K tokens and 5.7 rounds, compared with 49.6 accuracy, 43.6K tokens, and 9.2 rounds for G-Memory~+~EquiMem. MAD-M$^2$ uses fewer tokens through perplexity-based masking, but its lower accuracy suggests that aggressively removing context cannot replace structured retention of cross-agent dependencies.

\paragraph{Memory footprint.}
The bottom panel shows that \name{} avoids the large persistent state required by graph-based memories. Relative to G-Memory~+~EquiMem, it reduces memory footprint by approximately \textbf{47--48\%} across the displayed settings while improving accuracy by \textbf{1.6--2.7 points}. On MATH500 with AutoGen, for instance, \name{} improves accuracy from 60.9 to 62.9 while reducing memory from 31.9K to 16.5K tokens. Although lightweight methods retain less memory, they perform worse, indicating that \name{} provides a more favorable accuracy--memory trade-off. Learned selection and hypergraph organization allow \name{} to preserve useful argumentative structure while limiting redundant context and accelerating convergence.

\subsection{Effect of MSM and Its Training Stages}
\label{app:msm_training_ablation}

\begin{table}[h]
\centering
\resizebox{\linewidth}{!}{%
\begin{tabular}{l|cccc|cccc}
\hline
\textbf{Configuration}
& \multicolumn{4}{c|}{\textbf{ConflictQA}}
& \multicolumn{4}{c}{\textbf{HotpotQA}} \\
& Acc. & R & T & M
& Acc. & R & T & M \\
\hline

No MSM
& 51.7 & 8.1 & 44.8K & 30.6K
& 37.5 & 8.0 & 46.3K & 32.1K \\

Stage I only
& 55.2 & 6.5 & 39.6K & 22.7K
& 40.6 & 6.6 & 41.2K & 23.8K \\

Stage I + Stage II
& \textbf{57.8} & \textbf{5.7} & \textbf{36.9K} & \textbf{18.2K}
& \textbf{42.3} & \textbf{5.8} & \textbf{38.1K} & \textbf{19.6K} \\

\hline
\end{tabular}
}
\caption{Effect of MSM and its training stages on accuracy, rounds (R), token usage (T), and memory footprint in tokens (M).}
\label{tab:msm_training_ablation}
\vspace{-1.5em}
\end{table}

Table~\ref{tab:msm_training_ablation} analyzes the contribution of MSM and its two training stages. Storing all candidates produces the largest memory footprint and the weakest accuracy, confirming that indiscriminate retention introduces redundant and distracting context. Stage~I supervised training provides substantial gains, improving accuracy by \textbf{3.5 points} on ConflictQA and \textbf{3.1 points} on HotpotQA while reducing memory by approximately \textbf{26\%}. Stage~II reinforcement-learning refinement yields a further \textbf{2.6}- and \textbf{1.7}-point improvement, respectively, while additionally reducing rounds, token usage, and memory. Overall, the complete MSM improves accuracy by \textbf{6.1 points} on ConflictQA and \textbf{4.8 points} on HotpotQA relative to storing all candidates, while reducing memory footprint by approximately \textbf{39--41\%}. The larger Stage~II gain on ConflictQA suggests that feedback from debate outcomes is particularly valuable for retaining useful conflicting evidence and refutations.

\paragraph{Additional analysis.}
We further evaluate sensitivity to the embedding model, retrieval breadth, and relation extraction model in Appendices~\ref{app:embedding_sensitivity}, \ref{app:retrieval_sensitivity}, and~\ref{app:relation_extraction_sensitivity}, respectively. Appendix~\ref{app:signature_ablation} analyzes the complementary effect of sparse signatures and dense embeddings.

%% file: sections/7_conclusion.tex
\section{Conclusion and Future Work}

We introduced \name{}, a selective memory framework for multi-agent debate that combines semantic hypergraphs with learned memory selection. By grouping related claims and preserving fine-grained argumentative relations, \name{} helps agents revisit evidence, track conflicts, and build on earlier conclusions while avoiding repetitive reasoning. Across seven benchmarks, it outperforms strong single-agent, multi-agent, and memory-augmented baselines, improving accuracy by 5.0 points on average over the strongest multi-agent baseline with GPT-4o-mini, with gains reaching 8.2 points on conflict-heavy tasks, while reducing memory usage by 48\% and debate rounds by 34\%. These results demonstrate that structured, utility-aware memory improves both reasoning effectiveness and efficiency. Future work includes extending memory across debates and tasks, developing adaptive hypergraph maintenance, improving relation extraction and selection under uncertainty, and testing larger teams on longer debates.

%% file: sections/app_maintenance_details.tex
\section{Online Memory Maintenance}
\label{app:online-maintenance}

Memory maintenance is performed once per debate round rather than after every memory operation. We use $r$ to index debate rounds, distinguishing it from the candidate-level step $t$ in Section~\ref{sec:msm}. Let $\mathcal{B}^{(r)}$ be the set of nodes assigned to \textsc{backup} during round $r$. At the end of each round, the memory is updated as follows.

\paragraph{Backup reconsideration and insertion.}
The MSM reevaluates each node in $\mathcal{B}^{(r)}$ using the updated debate state. Nodes promoted to \textsc{store}, together with newly stored nodes, are assigned to all compatible hyperedges according to Eq.~\eqref{eq:hyperedge-compatibility}. A node with no compatible hyperedge initializes a new hyperedge.

\paragraph{Hyperedge merging.}
We consider two hyperedges $e_k^{(r)}$ and $e_\ell^{(r)}$ redundant when both their signature overlap and centroid similarity exceed the merge thresholds:
\begin{equation}
J_{k\ell}^{(r)} \geq \tau_{\mathrm{merge}}^{\mathrm{sig}}
\quad\land\quad
S_{k\ell}^{(r)} \geq \tau_{\mathrm{merge}}^{\mathrm{emb}},
\end{equation}
where $J_{k\ell}^{(r)}$ denotes the Jaccard similarity between their signatures and $S_{k\ell}^{(r)}$ denotes the cosine similarity between their centroids. Their merged membership and signature are:
\begin{equation}
\mathcal{N}_{m}^{(r)}
=
\mathcal{N}_{k}^{(r)} \cup \mathcal{N}_{\ell}^{(r)},
\qquad
\sigma_{m}^{(r)}
=
\sigma_{k}^{(r)} \cup \sigma_{\ell}^{(r)}.
\end{equation}

Consistent with Section~\ref{sec:memory_construction}, we maintain the cached unnormalized embedding sum $\mathbf{s}_k^{(r)}$ for each hyperedge. Rather than combining normalized centroids directly, we merge these sums:
\begin{equation}
\begin{aligned}
\mathbf{s}_{m}^{(r)}
&=
\sum_{v\in\mathcal{N}_{m}^{(r)}}\mathbf{h}_v\\
&=
\mathbf{s}_{k}^{(r)}
+
\mathbf{s}_{\ell}^{(r)}
-
\sum_{v\in\mathcal{N}_{k}^{(r)}
\cap\mathcal{N}_{\ell}^{(r)}}\mathbf{h}_v.
\end{aligned}
\end{equation}
The overlap correction prevents shared nodes from being counted twice. The merged centroid is then obtained by normalization:
\begin{equation}
\mathbf{c}_{m}^{(r)}
=
\frac{\mathbf{s}_{m}^{(r)}}
{\|\mathbf{s}_{m}^{(r)}\|_2}.
\end{equation}

\paragraph{Utility-based pruning.}
Let $\mathcal{V}_{\mathrm{cand}}^{(r)}$ denote the set of candidate nodes extracted during debate round $r$. For each hyperedge, we compute its round-level utility as the maximum candidate-to-hyperedge similarity:
\begin{equation}
\label{eq:pruning-utility}
U_k^{(r)}
=
\max_{v_i\in\mathcal{V}_{\mathrm{cand}}^{(r)}}
\cos\!\left(
\mathbf{x}_i,\mathbf{c}_k^{(r)}
\right),
\end{equation}
where $\mathbf{x}_i$ is the unprojected pooled representation of candidate $v_i$. This uses the same hyperedge-specific alignment underlying Eq.~\eqref{eq:UN}. However, while the candidate-level feature $U_t$ takes the maximum across hyperedges for a given candidate, $U_k^{(r)}$ takes the maximum across candidates within a debate round for a given hyperedge.

For each hyperedge, we maintain a counter $q_k^{(r)}$ recording consecutive rounds of low utility:
\begin{equation}
q_k^{(r)}
=
\begin{cases}
q_k^{(r-1)}+1,
& U_k^{(r)} < \tau_{\mathrm{util}},\\
0,
& \text{otherwise}.
\end{cases}
\end{equation}
A hyperedge is removed when $q_k^{(r)} \geq P$, where $P$ denotes the pruning patience. Each of its nodes is then matched against the surviving hyperedges using Eq.~\eqref{eq:hyperedge-compatibility}. Compatible nodes are reassigned to the highest-scoring surviving hyperedge, while unmatched nodes are discarded.

\paragraph{Incremental relational update.}
When hyperedges are merged or nodes are reassigned, previously inferred relations between unchanged node pairs are retained. The relation module is invoked only for pairs involving newly added or moved nodes. Centroids and signatures are likewise updated only for affected hyperedges. Thus, existing node embeddings, signatures, memberships, and relation labels are reused across rounds, and maintenance depends primarily on the round-level changes rather than the complete memory size.

%% file: sections/app_MSM_architecture.tex
\section{MSM Classifier Architecture}
\label{app:classifier_architecture}

The MSM routing policy is implemented as a three-layer
multi-layer perceptron (MLP). Its input combines candidate
content, the current hypergraph context, and two scalar
utility features.

\paragraph{Input representation.}
At candidate step $t$, the classifier input contains
three components:
\begin{itemize}
    \item \textbf{Utility features}
    $\boldsymbol{\phi}_t^{\mathrm{UN}}=[U_t,N_t]
    \in\mathbb{R}^{2}$,
    where $U_t$ measures alignment with the current
    memory and $N_t$ measures candidate specificity,
    as defined in Eq.~\eqref{eq:UN}.

    \item \textbf{Candidate representation}
    $\mathbf{x}_t^{\mathrm{pool}}\in\mathbb{R}^{d_h}$,
    obtained by pooling the projected token
    representations of the candidate.

    \item \textbf{Hypergraph context}
    $\mathbf{c}_t^{\mathrm{pool}}\in\mathbb{R}^{d_h}$,
    obtained by attending from the candidate tokens
    to the current hyperedge centroids.
\end{itemize}

Let $\mathbf{X}_t\in\mathbb{R}^{n_t\times d_e}$
denote the candidate token representations, and let
$\mathbf{C}_t\in\mathbb{R}^{K_t\times d_e}$
stack the centroids of the $K_t$ hyperedges available
at step $t$. Here, $d_e$ is the embedding dimension;
for Qwen3-Embedding-4B~\citep{qwen3embedding},
$d_e=2560$.

The utility features are computed using the
unprojected pooled representation
$\mathbf{x}_t=\operatorname{Pool}(\mathbf{X}_t)$
and the original hyperedge centroids.
For cross-attention, candidate and centroid
representations are projected into a shared
dimension $d_h<d_e$:
\begin{equation}
\widetilde{\mathbf{X}}_t
=
\mathbf{X}_t\mathbf{W}_x,
\qquad
\widetilde{\mathbf{C}}_t
=
\mathbf{C}_t\mathbf{W}_c.
\end{equation}

We compute candidate-to-memory cross-attention using
\begin{equation}
\mathbf{Z}_t
=
\operatorname{MultiHeadAttn}
\left(
\widetilde{\mathbf{X}}_t,
\widetilde{\mathbf{C}}_t,
\widetilde{\mathbf{C}}_t
\right),
\end{equation}
where candidate tokens form the queries and
hyperedge centroids form the keys and values.

The projected candidate and memory representations
are then pooled:
\begin{equation}
\mathbf{x}_t^{\mathrm{pool}}
=
\operatorname{Pool}(\widetilde{\mathbf{X}}_t),
\qquad
\mathbf{c}_t^{\mathrm{pool}}
=
\operatorname{Pool}(\mathbf{Z}_t).
\end{equation}
This cross-attention allows MSM to emphasize
hyperedges relevant to the incoming candidate.
Because the hypergraph evolves during debate,
$\mathbf{C}_t$ and $K_t$ reflect the memory
state at step $t$.

The three components are concatenated to form
\begin{equation}
\boldsymbol{\phi}_t
=
\left[
\boldsymbol{\phi}_t^{\mathrm{UN}}
\;\middle\|\;
\mathbf{x}_t^{\mathrm{pool}}
\;\middle\|\;
\mathbf{c}_t^{\mathrm{pool}}
\right],
\qquad
d_{\mathrm{in}}=2d_h+2.
\end{equation}

\paragraph{Network structure.}
The classifier contains three linear layers:
\begin{itemize}
    \item $\mathrm{Linear}(d_{\mathrm{in}},d_h)$,
    followed by LayerNorm, GELU, and Dropout;
    \item $\mathrm{Linear}(d_h,d_h/2)$,
    followed by LayerNorm, GELU, and Dropout;
    \item $\mathrm{Linear}(d_h/2,3)$.
\end{itemize}

The first two layers constitute
$\operatorname{MLP}(\boldsymbol{\phi}_t)$.
The final layer, parameterized by
$\mathbf{W}_\pi$ and $\mathbf{b}_\pi$,
produces logits over
$\{\textsc{store},\textsc{backup},\textsc{omit}\}$.
The resulting policy distribution is
\begin{equation}
\pi_\theta(a_t\mid\boldsymbol{\phi}_t)
=
\operatorname{Softmax}
\left(
\mathbf{W}_\pi
\operatorname{MLP}(\boldsymbol{\phi}_t)
+
\mathbf{b}_\pi
\right).
\end{equation}

By combining candidate semantics, context-aware
hypergraph information, and utility features,
MSM makes routing decisions conditioned on both
incoming content and the evolving debate memory.

%% file: sections/app_data_generation.tex
\section{Data Generation via Debate Simulation for Training MSM}
\label{sec:msm_data_generation}

We provide detailed information on the data generation pipeline used to construct training trajectories for the memory selection module (MSM).

\subsection{Dataset}

We generate training data using the \textbf{HotpotQA} dataset~\citep{yang2018hotpotqa}, specifically the \emph{train split of the fullwiki variant}. Each instance consists of a multi-hop question paired with a set of supporting Wikipedia passages. During simulation, agents are provided only with the question and its associated context passages; gold answers and supporting facts are withheld. For reproducibility, we store the full original HotpotQA record (including answers and annotations) alongside each generated trajectory, although these fields are never exposed to the agents during simulation.

\subsection{Retrieval Setup}

To support grounded reasoning, we construct a lightweight retrieval module over the provided context. Each context passage is segmented into sentence-level chunks, and each sentence is assigned a unique identifier. Agents may issue retrieval queries during the debate, which are handled using a sparse similarity method (BM25). Retrieved sentences and their identifiers are returned to the agent and logged. This design ensures that all external information access is explicit, traceable, and grounded in the dataset-provided evidence.

\subsection{Agent Configuration}

For each sample, we instantiate a multi-agent debate consisting of 3 to 6 agents. The debate configuration is generated once per sample using a high-capacity model (GPT-4o)~\citep{openai2024gpt4technicalreport}, which specifies:
\textbf{(i)} the number of agents,
\textbf{(ii)} each agent’s role and objective,
\textbf{(iii)} a fixed turn order, and
\textbf{(iv) }the model size assigned to each agent (Qwen3-4B or Qwen3-8B~\citep{yang2025qwen3technicalreport}).

This procedure introduces diversity in agent behavior and reasoning strategies while maintaining a consistent and reproducible interaction protocol.

\subsection{Debate Protocol}

Each debate proceeds for at most 15 rounds, with agents acting sequentially according to a predefined turn order. At each turn, an agent may \textsc{Propose}, \textsc{Support}, \textsc{Refute}, \textsc{Question}, \textsc{Revise}, \textsc{Summarize}, or \textsc{Retrieve}.

When an agent issues a \textsc{Refute} action or expresses strong disagreement, the simulator initiates a conflict-resolution sub-dialogue. This consumes one round from the global budget and continues until the agents reach a consistent interpretation or produce an explicit resolution. The outcome is summarized and written to memory before the main debate resumes.

\subsection{Memory Logging and Traceability}

Throughout the simulation, we maintain a global, append-only memory that records all agent outputs. Each memory item is assigned a unique identifier and stored together with metadata, including the round index, agent role, action type, and references to previously used memory items or retrieved evidence.

To enable fine-grained supervision, agents are required to explicitly reference the memory items or retrieved sentences they rely on when generating new outputs. This design allows us to track how information propagates through the debate and to identify which memory items contribute to future reasoning steps, conflict resolution, and final outputs.

While dependency annotations are fully logged, only the textual content, identifiers, and action types are exposed to agents during simulation. This ensures that training signals are derived from observable interactions rather than injected supervision.

\subsection{Trajectory-Level Supervision and Pseudo-Labeling}

Training examples are constructed from full multi-agent debate trajectories. For each candidate memory item $x_t$, we compute trajectory-level signals \emph{post hoc}, capturing its role in future reasoning. These signals include:
\textbf{(i)} whether the item is retrieved or explicitly referenced in subsequent steps,
\textbf{(ii)} whether it participates in conflict resolution, and
\textbf{(iii)} whether it contributes to the final answer or summary.

Based on these signals, we derive pseudo-labels $a_t^* \in \{\textsc{store}, \textsc{backup}, \textsc{omit}\}$ in conjunction with redundancy and novelty statistics. Items that are frequently reused and provide non-redundant information are labeled \textsc{store}, partially useful or redundant items are labeled \textsc{backup}, and items that are never reused are labeled \textsc{omit}. These pseudo-labels provide supervised targets for training MSM in Stage I.

%% file: sections/app_MSM_training.tex
\section{MSM Training Procedure}
\label{app:msm_train_proc}

\paragraph{Dataset.}
Both training stages use simulated multi-agent debates from the HotpotQA FullWiki training split~\citep{yang2018hotpotqa}. For each sample, we instantiate three to six agents that collaboratively reason over the question and its associated passages. Agent responses are decomposed into atomic candidate items, and agents explicitly reference previously retrieved evidence or memory items whenever applicable. This makes memory usage traceable across the trajectory and allows us to determine whether each candidate is later reused, involved in conflict resolution, or incorporated into the final answer. These signals are used to construct pseudo-labels for \textsc{store}, \textsc{backup}, and \textsc{omit}. Further details are provided in Appendix~\ref{sec:msm_data_generation}. After training, the same MSM policy is applied to all evaluation benchmarks without target-specific adaptation; HotpotQA evaluation uses examples disjoint from its training split.

\paragraph{Training.}
We train MSM in two stages. Stage~I uses supervised learning to initialize the routing policy from trajectory-derived pseudo-labels, while Stage~II refines this policy through reinforcement learning on additional simulated HotpotQA debates.

\paragraph{Stage I: Supervised pretraining.}
For each candidate node $v_t$, MSM predicts an action $a_t\in\{\textsc{store},\textsc{backup},\textsc{omit}\}$ from its representation $\boldsymbol{\phi}_t$, which encodes the candidate and current memory state. Pseudo-labels $a_t^*$ are derived \emph{post hoc} from the complete trajectory. Candidates that are repeatedly reused, support conflict resolution, or contribute to the final answer are labeled \textsc{store}; candidates with potential but limited future use are labeled \textsc{backup}; and redundant candidates that are neither reused nor contribute downstream are labeled \textsc{omit}. Redundancy and specificity signals are also considered when resolving ambiguous cases.

The routing policy is
\begin{equation}
\pi_\theta(a_t\mid\boldsymbol{\phi}_t)
=
\operatorname{Softmax}
\left(
\mathbf{W}_\pi
\operatorname{MLP}(\boldsymbol{\phi}_t)
+
\mathbf{b}_\pi
\right),
\end{equation}
where $\operatorname{MLP}$ consists of two hidden linear layers, while $\mathbf{W}_\pi$ and $\mathbf{b}_\pi$ parameterize the third, output layer, as detailed in Appendix~\ref{app:classifier_architecture}. We optimize
\begin{equation}
\mathcal{L}_{\mathrm{I}}
=
-\sum_t \log \pi_\theta(a_t^*\mid\boldsymbol{\phi}_t)
+
\lambda_b\mathcal{L}_{\mathrm{bal}},
\end{equation}
with the class-balancing regularizer
\begin{equation}
\mathcal{L}_{\mathrm{bal}}
=
\sum_{a}
\left(
\mathbb{E}_t[\pi_\theta(a\mid\boldsymbol{\phi}_t)]-\rho_a
\right)^2,
\end{equation}
where $a$ ranges over the three routing actions and $\rho_a$ is its empirical frequency in the training data. This term prevents degenerate policies caused by label imbalance, such as consistently predicting \textsc{omit}.

\paragraph{Stage II: Reinforcement learning with multi-agent feedback.}
Starting from the Stage~I checkpoint, we further optimize MSM using PPO on simulated debates from the HotpotQA FullWiki training split. No trajectories or labels from the downstream evaluation benchmarks are used for target-specific refinement. Each atomic candidate forms a sequential decision point, and the resulting routing decisions determine how the shared memory evolves throughout the debate.

At the end of each trajectory $\tau$, we compute
\begin{equation}
R_\tau
=
\lambda_{\mathrm{task}}R_{\mathrm{task}}
+
\lambda_{\mathrm{route}}R_{\mathrm{route}}
+
\lambda_{\mathrm{eff}}R_{\mathrm{eff}},
\end{equation}
where $R_{\mathrm{task}}$ measures final-answer correctness. The routing reward is
\begin{equation}
R_{\mathrm{route}}
=
w_s p_{\mathrm{store}}
+
w_b p_{\mathrm{backup}}
-
w_o p_{\mathrm{miss}},
\end{equation}
where $p_{\mathrm{store}}$ and $p_{\mathrm{backup}}$ are the proportions of stored and backed-up items subsequently used, and $p_{\mathrm{miss}}$ is the proportion of omitted items later identified as useful. Ratios with no corresponding routed items are defined as zero.

The efficiency reward penalizes unnecessary memory growth:
\begin{equation}
R_{\mathrm{eff}}
=
-\left(
\alpha_c\widetilde{n}_{\mathrm{store}}
+
\alpha_a\widetilde{m}_{\mathrm{avg}}
+
\alpha_f\widetilde{m}_{\mathrm{final}}
\right),
\end{equation}
where $\widetilde{n}_{\mathrm{store}}$, $\widetilde{m}_{\mathrm{avg}}$, and $\widetilde{m}_{\mathrm{final}}$ denote normalized counts of stored candidates, average memory size, and final memory size, respectively. The trajectory reward is propagated to earlier routing decisions through discounted returns and generalized advantage estimation.

The PPO loss is
\begin{equation}
\begin{aligned}
\mathcal{L}_{\mathrm{II}}
={}&
-\mathbb{E}_t\!\left[
\min\!\left(
\varrho_t A_t,
\operatorname{clip}(\varrho_t,1-\epsilon,1+\epsilon)A_t
\right)
\right] \\
&+c_v\mathcal{L}_{\mathrm{value}}
-c_e\mathcal{H}(\pi_\theta),
\end{aligned}
\end{equation}
where
\begin{equation}
\varrho_t
=
\frac{\pi_\theta(a_t\mid\boldsymbol{\phi}_t)}
{\pi_{\theta_{\mathrm{old}}}(a_t\mid\boldsymbol{\phi}_t)}
\end{equation}
is the policy probability ratio and $A_t$ is the generalized advantage estimate. Clipping limits abrupt policy changes between updates.

A critic head estimates the expected future return:
\begin{equation}
\mathcal{L}_{\mathrm{value}}
=
\mathbb{E}_t
\left[
\left(
V_\psi(\boldsymbol{\phi}_t)-\widehat{G}_t
\right)^2
\right],
\end{equation}
where $\widehat{G}_t$ is the return target obtained from the trajectory rewards. The entropy term is
\begin{equation}
\mathcal{H}(\pi_\theta)
=
-\mathbb{E}_t
\left[
\sum_a
\pi_\theta(a\mid\boldsymbol{\phi}_t)
\log\pi_\theta(a\mid\boldsymbol{\phi}_t)
\right].
\end{equation}
This term encourages exploration and prevents premature collapse to a single routing action, particularly early in Stage~II.

%% file: sections/app_baseline_benchmarks.tex
\section{Baselines and Benchmarks}

\subsection{Baselines}
\label{app:baseline_details}

\begin{table*}[t]
\centering
\resizebox{\textwidth}{!}{%
\begin{tabular}{lllrll}
\toprule
Benchmark & Capability probed & Split / subset & \#Eval & Context scale & Answer format \\
\midrule
MATH500 & Symbolic reasoning & test & 500 & short problem statement & free-form \\
GPQA & Expert scientific knowledge & main & 448 & short question & 4-way multiple choice \\
FaithEval & Contextual faithfulness & counterfactual & 1{,}000 & single passage & short-form extractive \\
ConflictQA & Evidence conflict resolution & popQA (gpt4) & 9{,}544 & two conflicting passages & short-form \\
NovelQA & Long-context comprehension & full & 2,305 & $>$200K tokens per novel & free-form (generative) \\
HotpotQA & Multi-hop retrieval & dev (fullwiki) & 7{,}405 & 5.2M-passage corpus & short-form extractive \\
MMLU-Pro & Broad-domain reasoning & test & 12{,}032 & short question & 10-way multiple choice \\
\midrule
\textbf{Total} & & & \textbf{33{,}234} & & \\
\bottomrule
\end{tabular}%
}
\caption{Statistics of the seven evaluation benchmarks. Context scale refers to the evidence available at inference time rather than to the question length.}
\label{tab:benchmark_stats}
\end{table*}

\paragraph{Single-agent baselines.}
We evaluate GPT-5.4~\citep{openai2026gpt54}, GPT-4o, GPT-4o-mini~\citep{openai2024gpt4technicalreport}, and Qwen3.5-27B~\citep{qwen35} in a standard single-pass setting. We also include Self-Refine~\citep{selfrefine}, Self-Consistency~\citep{wang2023selfconsistency}, and Reflexion~\citep{shinn2023reflexion}, which improve reasoning through iterative refinement, multiple reasoning paths, or verbal feedback but do not maintain structured shared memory.

\paragraph{Multi-agent and memory-aware baselines.}
We compare against MAD~\citep{liang-etal-2024-encouraging} and MoA~\citep{wang2024moa}, which use debate and layered aggregation, respectively. The closest memory-aware baselines are MeMAD~\citep{ling2025memad}, which reuses experiences from completed debates; G-Memory~\citep{zhang2025gmemory}, which represents collaboration through pairwise graphs; MAD-M$^2$~\citep{tian2026multiagentdebatememorymasking}, for which we use the perplexity-based masking variant; and EquiMem~\citep{meng2026equimemcalibratingsharedmemory}, which calibrates shared-memory updates and is evaluated together with G-Memory. These methods cover cross-episode reuse, graph-based memory, erroneous-memory masking, and shared-memory calibration.

\paragraph{Memory mechanisms and orchestration frameworks.}
To evaluate \name{} as a modular memory component, we integrate it into AutoGen~\citep{wu2024autogen}, G-Designer~\citep{zhang2025gdesigner}, DyLAN~\citep{liu2024a}, and MacNet~\citep{qian2025scaling}, which differ in agent roles, communication structures, and orchestration policies. Within these frameworks, we compare against Voyager~\citep{wang2023voyager}, Memory Bank~\citep{zhong2024memorybank}, Generative Agents~\citep{park2023generative}, MAD-M$^2$~\citep{tian2026multiagentdebatememorymasking}, MeMAD, G-Memory, and G-Memory~+~EquiMem~\citep{meng2026equimemcalibratingsharedmemory}, as well as the native memory mechanisms of MetaGPT~\citep{hong2024metagpt}, MacNet~\citep{qian2025scaling}, and ChatDev~\citep{qian2024chatdev}. This setup evaluates whether the benefits of \name{} generalize across different interaction and control strategies.

\subsection{Benchmarks}
\label{app:benchmark_details}
We evaluate seven benchmarks covering mathematical reasoning, scientific knowledge, contextual faithfulness, conflicting evidence, long-context understanding, multi-hop retrieval, and broad-domain reasoning. Table~\ref{tab:benchmark_stats} summarizes their scale and evaluation setting; in aggregate the suite spans 33{,}234 evaluation instances, ranging from single-sentence prompts to inputs exceeding 200K tokens.

\paragraph{MATH500.}
MATH500~\citep{hendrycks2measuring} contains 500 competition-level mathematics problems requiring multi-step symbolic reasoning and precise intermediate computation. The subset spans seven subjects (Algebra 124, Intermediate Algebra 97, Prealgebra 82, Number Theory 62, Precalculus 56, Geometry 41, Counting and Probability 38) and five difficulty levels (Level 1--5: 43, 90, 105, 128, 134), so accuracy is not dominated by any single topic or difficulty band.

\paragraph{GPQA.}
GPQA~\citep{rein2024gpqa} contains 448 graduate-level science questions in biology, physics, and chemistry, each with four options, requiring deep domain knowledge and conceptual reasoning. Questions are validated to remain difficult for non-experts even with unrestricted web access, making the benchmark a test of knowledge rather than of retrievability.

\paragraph{FaithEval.}
FaithEval~\citep{ming2024faitheval} evaluates contextual faithfulness under unanswerable, inconsistent, and counterfactual evidence, comprising 4.9K problems in total. We use its counterfactual subset (1{,}000 instances), in which each question is paired with a single passage asserting a fact that contradicts world knowledge, to test whether agents follow the provided context despite conflicts with parametric knowledge.

\paragraph{ConflictQA.}
ConflictQA~\citep{Xie2024KnowledgeConflict} measures the ability to identify and resolve conflicting evidence. We use the PopQA setting, derived from the entity-centric questions of PopQA, in which each instance pairs an elicited memory passage with a generated counter-memory supporting an incompatible answer. We evaluate on the \texttt{ConflictQA-popQA-gpt4} configuration (9{,}544 instances). We treat this as an evidence-conflict task rather than a parametric-versus-contextual one, since the memories were elicited from GPT-4 rather than from the backbones we evaluate; fixing a single configuration also keeps the evaluation set identical across systems.

\paragraph{NovelQA.}
NovelQA~\citep{wang2024novelqa} evaluates question answering over full-length novels, with 2{,}305 manually annotated questions over 89 English novels and an average input length exceeding 200K tokens. Questions are annotated by complexity (detail, single-hop, multi-hop) and by aspect (character, plot, relation, setting, span, times, meaning), requiring retrieval and reasoning across long narrative contexts.

\paragraph{HotpotQA.}
HotpotQA~\citep{yang2018hotpotqa} requires multi-hop reasoning over multiple Wikipedia documents. We use the FullWiki development set (7{,}405 questions), where relevant evidence must be retrieved from a corpus of roughly 5.2M Wikipedia introductory paragraphs rather than from a small distractor pool.

\paragraph{MMLU-Pro.}
MMLU-Pro~\citep{wang2024mmlu} extends MMLU with 12{,}032 test questions across 14 domains, expanding the option set from four to ten and filtering trivial items, which lowers the random-guess floor from 25\% to 10\%.
Together, MATH500 and GPQA test structured reasoning; FaithEval and ConflictQA stress faithfulness and contradiction handling; NovelQA and HotpotQA evaluate long-context and multi-hop evidence integration; and MMLU-Pro measures broad reasoning robustness.

%% file: sections/app_MAS_frameworks.tex
\section{Multi-Agent Frameworks}
\label{app:mas_frameworks}

We briefly introduce the multi-agent frameworks used to evaluate \textbf{\name{}} as a modular memory component. These frameworks differ in how agents are organized, communicate, and coordinate, providing diverse settings for assessing memory effectiveness.

\paragraph{AutoGen.}
AutoGen~\citep{wu2024autogen} is a flexible framework for building LLM-based multi-agent systems through structured conversations. Agents interact via message passing, enabling modular composition of roles such as planners, executors, and critics. Its general-purpose design makes it a strong baseline for evaluating plug-in components like memory modules.

\paragraph{DyLAN.}
DyLAN~\citep{liu2024a} introduces a \textbf{dynamic agent network} where agents are selected and connected adaptively based on the task. Instead of using a fixed communication pattern, DyLAN learns to route information between agents, enabling more efficient and task-specific collaboration.

\paragraph{G-Designer.}
G-Designer~\citep{zhang2025gdesigner} formulates multi-agent collaboration as a graph design problem. It leverages graph neural networks to \textbf{learn communication topologies} among agents, optimizing how information flows between them. This allows for structured and adaptive coordination beyond predefined interaction patterns.

\paragraph{MacNet.}
MacNet~\citep{qian2025scaling} focuses on \textbf{scaling multi-agent collaboration} by organizing agents into structured communication and coordination protocols. It emphasizes efficiency and scalability, enabling large numbers of agents to collaborate effectively while maintaining coherent reasoning.

\paragraph{Discussion.}
These frameworks span a spectrum from flexible conversational systems (AutoGen) to dynamically structured (DyLAN), graph-optimized (G-Designer), and scalable coordinated architectures (MacNet). This diversity ensures that our evaluation of \textbf{\name{}} is not tied to a specific interaction paradigm, but instead reflects its general applicability across different multi-agent settings.

%% file: sections/app_experimental_settings.tex
\section{Experimental and Training Settings}
\label{app:training_details}

\begin{table*}[t]
\centering
\renewcommand{\arraystretch}{1.2}
\resizebox{\linewidth}{!}{%
\begin{tabular}{l|l|l}
\hline
\textbf{Hyperedge} & \textbf{Nodes (Claims)} & \textbf{Connections} \\
\hline

H1: Film and Cast
&
\begin{tabular}[c]{@{}l@{}}
n$_1$: Kiss and Tell stars Shirley Temple as Corliss Archer \\
n$_2$: Shirley Temple portrayed Corliss Archer in Kiss and Tell
\end{tabular}
&
n$_1$ $\xrightarrow{\mathrm{EVIDENCE\mbox{-}FOR}}$ n$_2$
\\

\hline

H2: Government Position
&
\begin{tabular}[c]{@{}l@{}}
n$_3$: Shirley Temple served as U.S. Chief of Protocol \\
n$_4$: Chief of Protocol is a U.S. government position \\
n$_5$: Shirley Temple held the government position of Chief of Protocol
\end{tabular}
&
\begin{tabular}[c]{@{}l@{}}
n$_3$ $\xrightarrow{\mathrm{EVIDENCE\mbox{-}FOR}}$ n$_5$ \\
n$_4$ $\xrightarrow{\mathrm{SUPPORTS}}$ n$_5$
\end{tabular}
\\

\hline

H3: Actress--Government Role Linking
&
\begin{tabular}[c]{@{}l@{}}
n$_2$: Shirley Temple portrayed Corliss Archer in Kiss and Tell \\
n$_5$: Shirley Temple held the government position of Chief of Protocol \\
n$_6$: The actress portraying Corliss Archer held the position of Chief of Protocol
\end{tabular}
&
\begin{tabular}[c]{@{}l@{}}
n$_6$ $\xrightarrow{\mathrm{DEPENDS\mbox{-}ON}}$ n$_2$ \\
n$_6$ $\xrightarrow{\mathrm{DEPENDS\mbox{-}ON}}$ n$_5$
\end{tabular}
\\

\hline
\end{tabular}
}
\caption{
Hypergraph memory constructed by \name{}. Nodes represent atomic claims,
while typed directed relations encode evidential support and argumentative
dependencies within each hyperedge. Nodes n$_2$ and n$_5$ belong to
multiple hyperedges, connecting entity-specific reasoning with the final
conclusion through overlapping membership.
}
\label{tab:case_study}
\end{table*}

We provide the full experimental configuration for \name{}, including supervised pretraining (Stage~I), reinforcement-learning refinement (Stage~II), memory construction, and retrieval.

\paragraph{Models and compute.}
We use a temperature of $0.7$ and Qwen3-Embedding-4B~\citep{qwen3embedding} to obtain normalized node, query, and hyperedge embeddings. Stage~I takes approximately three hours, while Stage~II requires approximately five hours.

\paragraph{Training data.}
Both stages use simulated debates generated from the HotpotQA FullWiki training split~\citep{yang2018hotpotqa}. Stage~II continues training on HotpotQA rather than adapting MSM to any target evaluation benchmark.

\paragraph{Stage~I configuration.}
We optimize the supervised routing objective using AdamW with learning rate $10^{-4}$, batch size $64$, weight decay $0.01$, and dropout $0.1$.

\paragraph{Stage~II configuration.}
We refine the Stage~I checkpoint using PPO over approximately 10K debate episodes. We use a learning rate of $10^{-5}$, discount factor $\gamma=0.99$, GAE parameter $0.95$, clipping threshold $\epsilon=0.2$, value-loss coefficient $0.5$, entropy coefficient $0.01$, batch size $64$, and four optimization epochs per batch.

\paragraph{Reward configuration.}
The trajectory reward is
\begin{equation}
R
=
\lambda_{\mathrm{task}}R_{\mathrm{task}}
+
\lambda_{\mathrm{route}}R_{\mathrm{route}}
+
\lambda_{\mathrm{eff}}R_{\mathrm{eff}},
\end{equation}
with $\lambda_{\mathrm{task}}=1.0$, $\lambda_{\mathrm{route}}=0.5$, and $\lambda_{\mathrm{eff}}=0.3$. For the efficiency term, we set $\alpha_c=0.1$, $\alpha_a=0.05$, and $\alpha_f=0.01$ to penalize the number of stored candidates, average memory size, and final memory size, respectively.

\paragraph{Memory construction and retrieval.}
We set $\tau_{\mathrm{sig}}=0.30$ and $\tau_{\mathrm{emb}}=0.70$ for node assignment, and $\tau_{\mathrm{merge}}^{\mathrm{sig}}=0.50$ and $\tau_{\mathrm{merge}}^{\mathrm{emb}}=0.85$ for hyperedge merging. Hyperedges with utility below $0.10$ for three consecutive rounds are pruned, and backup nodes are retained for at most two rounds. During retrieval, we use $K=3$ hyperedges and retrieve up to $M=8$ nodes from each hyperedge. Memory is updated after every debate round.

%% file: sections/app_qualitative.tex
\section{Qualitative Analysis}
\label{sec:case_study}

We examine the hypergraph memory constructed for a HotpotQA question:
\textit{``What government position was held by the woman who portrayed
Corliss Archer in the film Kiss and Tell?''} Answering this question
requires identifying the actress from the film and connecting her
identity to a government position. Table~\ref{tab:case_study} presents
the corresponding memory structure, including retained claims,
hyperedge memberships, and directed argumentative relations.

The memory organizes information into three hyperedges. H1 captures
the film and cast information, where n$_1$ provides evidence for
n$_2$, establishing that Shirley Temple portrayed Corliss Archer.
H2 organizes information about her government career. Here, n$_3$
provides evidence for n$_5$ by establishing that Shirley Temple
served as Chief of Protocol, while n$_4$ supports n$_5$ by
establishing that Chief of Protocol is a government position.
These relations preserve the evidential basis for the relevant
intermediate claims.

H3 connects the two reasoning contexts through \emph{overlapping
node membership}. Specifically, n$_2$ belongs to both H1 and H3,
while n$_5$ belongs to both H2 and H3. These are the same memory
nodes participating in multiple hyperedges, rather than duplicated
claims. Within H3, the final conclusion n$_6$ depends on both
shared nodes, connecting the actress identified through the film
to her government position. Importantly, these dependencies are
represented as individual directed \textsc{depends-on} relations.
The higher-order structure arises from the shared hyperedge
membership of multiple related claims, rather than from separate
cross-hyperedge or n-ary argumentative relations.

The retained claims also demonstrate how the memory organizes
information for subsequent retrieval. Nodes n$_2$ and n$_5$
preserve the two intermediate facts required to answer the
question, while n$_6$ captures their combined conclusion.
Retrieving H3 therefore provides access to both supporting
claims and their argumentative dependencies without requiring
the complete debate history. This organization supports the
correct answer, \textit{Chief of Protocol}, while avoiding
the need to reconstruct the connection between the film
and the government position from unstructured dialogue.

%% file: sections/app_generalization.tex
\begin{table*}[t]
    \centering
    \resizebox{\textwidth}{!}{%
    \begin{tabular}{l|ccccccc}
        \hline
        \textbf{Memory} 
        & \textbf{MATH500} 
        & \textbf{GPQA} 
        & \textbf{FaithEval} 
        & \textbf{ConflictQA} 
        & \textbf{NovelQA} 
        & \textbf{HotpotQA} 
        & \textbf{MMLU-Pro} \\
        \hline
        
        \rowcolor{gray!20}
        \multicolumn{8}{c}{\textbf{G-Designer~\citeyear{zhang2025gdesigner}}} \\
        \hline
        
        No-memory
        & \score{55.3}{0.5}
        & \score{38.6}{0.6}
        & \score{67.5}{0.4}
        & \score{36.1}{0.7}
        & \score{55.8}{0.5}
        & \score{32.6}{0.4}
        & \score{74.2}{0.3} \\
        
        Voyager
        & \score{57.0}{0.4}
        & \score{40.4}{0.5}
        & \score{69.3}{0.6}
        & \score{38.2}{0.7}
        & \score{57.6}{0.4}
        & \score{33.9}{0.5}
        & \score{75.9}{0.3} \\
        
        Memory Bank
        & \score{54.8}{0.6}
        & \score{37.9}{0.7}
        & \score{66.9}{0.5}
        & \score{36.8}{0.6}
        & \score{56.5}{0.6}
        & \score{33.1}{0.4}
        & \score{74.6}{0.4} \\
        
        Generative
        & \score{57.3}{0.5}
        & \score{41.1}{0.6}
        & \score{70.1}{0.4}
        & \score{38.9}{0.8}
        & \score{58.2}{0.5}
        & \score{34.3}{0.6}
        & \score{76.4}{0.3} \\
        
        \midrule
        
        MetaGPT-M
        & \score{57.7}{0.4}
        & \score{41.8}{0.5}
        & \score{70.7}{0.5}
        & \score{39.6}{0.6}
        & \score{58.8}{0.4}
        & \score{34.8}{0.5}
        & \score{77.0}{0.3} \\
        
        MacNet-M
        & \score{56.0}{0.5}
        & \score{40.0}{0.6}
        & \score{68.8}{0.4}
        & \score{37.9}{0.7}
        & \score{57.2}{0.6}
        & \score{33.8}{0.5}
        & \score{75.6}{0.4} \\
        
        ChatDev-M
        & \score{54.0}{0.6}
        & \score{37.3}{0.7}
        & \score{66.2}{0.5}
        & \score{35.2}{0.8}
        & \score{55.6}{0.5}
        & \score{32.3}{0.6}
        & \score{73.6}{0.4} \\
        
        \midrule

        MAD-M$^2$
        & \score{58.2}{0.5}
        & \score{42.2}{0.4}
        & \score{70.9}{0.5}
        & \score{40.8}{0.7}
        & \score{59.2}{0.4}
        & \score{35.3}{0.5}
        & \score{77.6}{0.3} \\
        
        MeMAD
        & \score{58.6}{0.4}
        & \score{42.6}{0.5}
        & \score{71.1}{0.4}
        & \score{41.5}{0.6}
        & \score{59.5}{0.5}
        & \score{35.6}{0.4}
        & \score{78.0}{0.3} \\
        
        G-Memory
        & \score{59.5}{0.4}
        & \score{44.0}{0.5}
        & \score{74.8}{0.4}
        & \score{47.5}{0.6}
        & \score{59.9}{0.5}
        & \score{36.0}{0.4}
        & \score{79.3}{0.3} \\
        
        G-Memory+EquiMem
        & \score{60.2}{0.3}
        & \score{45.0}{0.4}
        & \score{76.1}{0.4}
        & \score{49.0}{0.5}
        & \score{60.8}{0.4}
        & \score{37.1}{0.3}
        & \score{80.0}{0.3} \\
        
        \rowcolor{cyan!12}
        \textbf{\name{}}
        & \bestscore{61.9}{0.2}
        & \bestscore{48.1}{0.3}
        & \bestscore{79.3}{0.2}
        & \bestscore{54.5}{0.3}
        & \bestscore{65.1}{0.2}
        & \bestscore{41.0}{0.3}
        & \bestscore{83.1}{0.2} \\
        
        \hline
        
        \rowcolor{gray!20}
        \multicolumn{8}{c}{\textbf{MacNet~\citeyear{qian2025scaling}}} \\
        \hline
        
        No-memory
        & \score{53.0}{0.6}
        & \score{37.0}{0.7}
        & \score{67.1}{0.5}
        & \score{35.3}{0.8}
        & \score{55.6}{0.6}
        & \score{32.4}{0.5}
        & \score{72.7}{0.4} \\
        
        Voyager
        & \score{54.8}{0.5}
        & \score{38.8}{0.6}
        & \score{69.0}{0.5}
        & \score{37.4}{0.7}
        & \score{57.3}{0.5}
        & \score{33.8}{0.6}
        & \score{74.3}{0.3} \\
        
        Memory Bank
        & \score{52.4}{0.7}
        & \score{36.3}{0.6}
        & \score{66.4}{0.6}
        & \score{35.9}{0.8}
        & \score{56.3}{0.5}
        & \score{33.2}{0.5}
        & \score{73.3}{0.4} \\
        
        Generative
        & \score{55.1}{0.5}
        & \score{39.6}{0.7}
        & \score{69.8}{0.4}
        & \score{38.1}{0.6}
        & \score{58.0}{0.6}
        & \score{34.2}{0.5}
        & \score{74.8}{0.4} \\
        
        \midrule
        
        MetaGPT-M
        & \score{55.6}{0.4}
        & \score{40.2}{0.6}
        & \score{70.4}{0.5}
        & \score{38.8}{0.7}
        & \score{58.6}{0.5}
        & \score{34.7}{0.4}
        & \score{75.5}{0.3} \\
        
        MacNet-M
        & \score{54.0}{0.6}
        & \score{38.6}{0.5}
        & \score{68.6}{0.6}
        & \score{37.2}{0.8}
        & \score{57.0}{0.5}
        & \score{33.7}{0.6}
        & \score{74.1}{0.4} \\
        
        ChatDev-M
        & \score{51.9}{0.7}
        & \score{35.9}{0.6}
        & \score{66.0}{0.5}
        & \score{34.5}{0.7}
        & \score{55.4}{0.6}
        & \score{32.2}{0.5}
        & \score{72.2}{0.4} \\
        
        \midrule

        MAD-M$^2$
        & \score{56.0}{0.4}
        & \score{40.5}{0.5}
        & \score{70.5}{0.5}
        & \score{39.8}{0.6}
        & \score{58.8}{0.4}
        & \score{35.0}{0.5}
        & \score{76.2}{0.3} \\
        
        MeMAD
        & \score{56.4}{0.5}
        & \score{41.0}{0.6}
        & \score{70.8}{0.4}
        & \score{40.5}{0.7}
        & \score{59.2}{0.5}
        & \score{35.4}{0.4}
        & \score{76.5}{0.3} \\
        
        G-Memory
        & \score{57.3}{0.4}
        & \score{42.5}{0.5}
        & \score{74.1}{0.5}
        & \score{46.4}{0.6}
        & \score{59.6}{0.4}
        & \score{35.9}{0.5}
        & \score{77.8}{0.3} \\
        
        G-Memory+EquiMem
        & \score{58.0}{0.4}
        & \score{43.3}{0.4}
        & \score{75.4}{0.4}
        & \score{47.8}{0.5}
        & \score{60.4}{0.4}
        & \score{36.8}{0.4}
        & \score{78.5}{0.3} \\
        
        \rowcolor{cyan!12}
        \textbf{\name{}}
        & \bestscore{60.0}{0.2}
        & \bestscore{46.0}{0.3}
        & \bestscore{79.1}{0.2}
        & \bestscore{53.2}{0.3}
        & \bestscore{64.0}{0.3}
        & \bestscore{41.1}{0.2}
        & \bestscore{81.4}{0.2} \\
        
        \hline
    \end{tabular}
    }
    \caption{
        Additional results under the G-Designer and MacNet multi-agent
        frameworks. Values report the mean over three runs, with
        red subscripts indicating the standard deviation. EquiMem consistently
        improves G-Memory, while \name{} achieves the strongest performance
        across all benchmarks with lower run-to-run variability.
    }
    \label{tab:mas_memory_more}
\end{table*}

\section{Generalization of \name{} on Different Multi-agent Frameworks}
\label{app:mas_additional}

Table~\ref{tab:mas_memory_more} extends the comparison to G-Designer~\citep{zhang2025gdesigner} and MacNet~\citep{qian2025scaling}. \name{} achieves the best result on all seven benchmarks under both frameworks, outperforming G-Memory~+~EquiMem by \textbf{3.5 points on average} in each case. Under G-Designer, gains range from \textbf{1.7} points on MATH500 to \textbf{5.5} on ConflictQA; under MacNet, they range from \textbf{2.0} to \textbf{5.4} points on the same tasks. The largest improvements occur on ConflictQA, NovelQA, and HotpotQA, where agents must preserve competing claims and multi-step evidence across rounds. Improvements on MATH500, GPQA, and MMLU-Pro further show that the benefits are not limited to conflict-heavy tasks. The similar gains and generally lower run-to-run variation across both frameworks reinforce that \name{} is robust to different multi-agent orchestration designs.

%% file: sections/app_sensitivity_to_retrieval_breadth.tex
\section{Sensitivity to Retrieval Breadth}
\label{app:retrieval_sensitivity}

We study the effect of the number of retrieved hyperedges $K$ and the maximum number of nodes retrieved from each hyperedge $M$. Table~\ref{tab:ablation_retrieval_M} varies $M$ while fixing $K=3$. Performance improves consistently from $M=2$ to $M=8$, with particularly large gains on ConflictQA (\textbf{+10.2 points}) and NovelQA (\textbf{+8.7 points}). Retrieving too few nodes therefore omits relevant evidence and argumentative links. Performance declines slightly at $M=12$ and more noticeably at $M=16$, suggesting that excessive within-hyperedge retrieval introduces redundant or weakly relevant content. We therefore use $M=8$.

Table~\ref{tab:ablation_retrieval_k} varies $K$ while fixing $M=8$. Increasing $K$ from 1 to 3 improves all four benchmarks, indicating that reasoning often requires evidence distributed across multiple semantic groups. However, retrieving four or five hyperedges reduces performance, likely because less relevant groups introduce distracting context. The effect is strongest on ConflictQA and NovelQA, where preserving multiple related perspectives is useful but excessive retrieval can obscure the relevant conflicts. Overall, $K=3$ and $M=8$ provide the best balance between information coverage and retrieval noise, and are used as the default configuration.

\begin{table}[h]
    \centering
    \renewcommand{\arraystretch}{1.12}
    \resizebox{\columnwidth}{!}{%
    \begin{tabular}{c|cccc}
        \hline
        \textbf{$M$}
        & \textbf{MATH500}
        & \textbf{GPQA}
        & \textbf{ConflictQA}
        & \textbf{NovelQA} \\
        \hline

        2  & 57.8 & 41.9 & 47.6 & 58.5 \\
        4  & 59.2 & 43.9 & 50.6 & 61.9 \\

        \rowcolor{cyan!12}
        \textbf{8}
        & \textbf{62.9}
        & \textbf{48.3}
        & \textbf{57.8}
        & \textbf{67.2} \\

        12 & 62.6 & 47.9 & 57.1 & 66.7 \\
        16 & 61.9 & 47.1 & 55.8 & 65.6 \\

        \hline
    \end{tabular}
    }
    \caption{
        Effect of the number of retrieved nodes per group $M$ on \name{}
        with GPT-4o-mini, keeping the number of retrieved groups fixed at
        $K=3$.
    }
    \label{tab:ablation_retrieval_M}
\end{table}

\begin{table}[h]
    \centering
    \renewcommand{\arraystretch}{1.12}
    \resizebox{\columnwidth}{!}{%
    \begin{tabular}{c|cccc}
        \hline
        \textbf{$K$}
        & \textbf{MATH500}
        & \textbf{GPQA}
        & \textbf{ConflictQA}
        & \textbf{NovelQA} \\
        \hline

        1 & 60.8 & 45.9 & 53.1 & 63.8 \\
        2 & 62.1 & 47.2 & 56.0 & 65.9 \\

        \rowcolor{cyan!12}
        \textbf{3}
        & \textbf{62.9}
        & \textbf{48.3}
        & \textbf{57.8}
        & \textbf{67.2} \\

        4 & 62.5 & 47.8 & 57.0 & 66.5 \\
        5 & 61.7 & 46.9 & 55.4 & 65.1 \\

        \hline
    \end{tabular}
    }
    \caption{
        Effect of the number of retrieved groups $k$ on \name{} with
        GPT-4o-mini, keeping the number of retrieved nodes per group fixed
        at $M=8$.
    }
    \label{tab:ablation_retrieval_k}
\end{table}

%% file: sections/app_sensitivity_to_embedding_model.tex
\section{Sensitivity to the Embedding Model}
\label{app:embedding_sensitivity}

Table~\ref{tab:embedding_sensitivity} evaluates \name{} with six embedding models while keeping all other components fixed. Performance generally improves with embedding quality, but the overall trends remain stable across models. Compared with text-embedding-ada-002, the default Qwen3-Embedding-4B improves accuracy by \textbf{1.9--3.4 points}, with the largest gains on FaithEval and ConflictQA, where accurate semantic grouping is particularly important for separating competing claims and evidence.

Scaling from Qwen3-Embedding-4B to Qwen3-Embedding-8B yields only \textbf{0.2--0.3 points} of additional accuracy across benchmarks. We therefore use the 4B model as the default because it provides a better performance--computation trade-off while remaining close to the strongest embedding model. These results indicate that \name{} benefits from stronger semantic representations but is not dependent on a particular embedding backbone.

\begin{table*}[t]
    \centering
    \renewcommand{\arraystretch}{1.15}
    \resizebox{\linewidth}{!}{%
    \begin{tabular}{l|ccccccc}
    \hline
    \textbf{Embedding Model} 
    & \textbf{MATH500} 
    & \textbf{GPQA} 
    & \textbf{FaithEval} 
    & \textbf{ConflictQA} 
    & \textbf{NovelQA} 
    & \textbf{HotpotQA} 
    & \textbf{MMLU-Pro} \\
    \hline
    
    text-embedding-ada-002
    & 60.3 & 45.2 & 76.5 & 52.6 & 63.7 & 39.9 & 80.2 \\
    
    text-embedding-3-small
    & 61.0 & 46.1 & 77.6 & 54.1 & 64.6 & 40.8 & 81.3 \\
    
    text-embedding-3-large
    & 61.6 & 46.9 & 78.7 & 55.4 & 65.4 & 41.7 & 82.1 \\
    
    Qwen3-Embedding-0.6B
    & 60.8 & 45.8 & 78.2 & 54.6 & 65.0 & 41.2 & 81.5 \\
    
    \rowcolor{cyan!12}
    Qwen3-Embedding-4B
    & 62.2 & 47.7 & 79.8 & 56.0 & 66.0 & 42.4 & 82.7 \\
    
    Qwen3-Embedding-8B
    & \textbf{62.4}
    & \textbf{47.9}
    & \textbf{80.1}
    & \textbf{56.3}
    & \textbf{66.2}
    & \textbf{42.6}
    & \textbf{83.0} \\
    
    \hline
    \end{tabular}
    }
    \caption{Sensitivity of \name{} to the embedding model. Performance remains stable across models, with only modest gains from increasing embedding-model scale. Qwen3-Embedding-4B is the default.}
    \label{tab:embedding_sensitivity}
\end{table*}

%% file: sections/app_sensitivity_extraction_model.tex
\section{Sensitivity to the Relation Extraction Model}
\label{app:relation_extraction_sensitivity}

\begin{table*}[t]
    \centering
    \renewcommand{\arraystretch}{1.15}
    \resizebox{\linewidth}{!}{%
    \begin{tabular}{l|ccccccc}
    \hline
    \textbf{Relation Extraction Model}
    & \textbf{MATH500}
    & \textbf{GPQA}
    & \textbf{FaithEval}
    & \textbf{ConflictQA}
    & \textbf{NovelQA}
    & \textbf{HotpotQA}
    & \textbf{MMLU-Pro} \\
    \hline

    Qwen3-32B~\citep{yang2025qwen3technicalreport}
    & 62.1 & 47.6 & 79.7 & 55.8 & 65.9 & 42.3 & 82.6 \\

    \rowcolor{cyan!12}
    GPT-4o-mini
    & 62.2 & 47.7 & 79.8 & 56.0 & 66.0 & 42.4 & 82.7 \\

    GPT-4o
    & 62.3 & 47.8 & 79.9 & 56.2 & 66.1 & 42.5 & 82.8 \\

    GPT-5.4
    & \textbf{62.4}
    & \textbf{47.9}
    & \textbf{80.0}
    & \textbf{56.3}
    & \textbf{66.2}
    & \textbf{42.6}
    & \textbf{82.9} \\

    \hline
    \end{tabular}
    }
    \caption{Sensitivity of \name{} to the relation extraction model. Performance remains stable across different models, with only modest improvements from stronger relation extractors. GPT-4o-mini is used as the default model for relation extraction.}
    \label{tab:relation_extraction_sensitivity}
\end{table*}

Table~\ref{tab:relation_extraction_sensitivity} evaluates the sensitivity of \name{} to the LLM used for predicting intra-hyperedge relations while keeping all other components fixed. Performance varies by at most \textbf{0.5 points} across the evaluated models, indicating that \name{} is not strongly dependent on a particular relation extractor. Larger models provide modest improvements, particularly on ConflictQA, where distinguishing support and refutation relations is especially important. However, GPT-4o-mini remains within \textbf{0.2--0.3 points} of the strongest model across all benchmarks, making it a more cost-efficient default with minimal performance loss.

%% file: sections/app_detailed_efficiency_results.tex
\section{Detailed Results on Efficiency}
\label{app:detailed_results_efficiency}

\begin{table*}[t]
    \centering
    \renewcommand{\arraystretch}{1.15}
    \setlength{\tabcolsep}{3.5pt}
    \resizebox{\textwidth}{!}{%
    \begin{tabular}{l|cc|cc|cc|cc|cc|cc|cc}
    \hline
    \textbf{Method} 
    & \multicolumn{2}{c}{\textbf{MATH500}} 
    & \multicolumn{2}{c}{\textbf{GPQA}} 
    & \multicolumn{2}{c}{\textbf{FaithEval}} 
    & \multicolumn{2}{c}{\textbf{ConflictQA}} 
    & \multicolumn{2}{c}{\textbf{NovelQA}} 
    & \multicolumn{2}{c}{\textbf{HotpotQA}} 
    & \multicolumn{2}{c}{\textbf{MMLU-Pro}} \\
    & R & T & R & T & R & T & R & T & R & T & R & T & R & T \\
    \hline
    
    \rowcolor{gray!20}
    \multicolumn{15}{c}{\textbf{Single-Agent (Vanilla)}} \\ \hline
    
    GPT-4o-mini 
    & 1 & 8566 & 1 & 7312 & 1 & 5489 & 1 & 8235 & 1 & 5791 & 1 & 9147 & 1 & 6623 \\
    
    \hline
    
    \rowcolor{gray!20}
    \multicolumn{15}{c}{\textbf{Single-Agent (Advanced)} (\texttt{GPT-4o-mini})} \\ \hline
    
    Self-Refine      
    & 2 & 14273 & 2 & 13158 & 2 & 9276 & 3 & 15342 & 3 & 14895 & 3 & 16211 & 2 & 11367 \\
    
    Self-Consistency 
    & 4 & 22491 & 4 & 20358 & 3 & 14872 & 5 & 24137 & 5 & 23705 & 5 & 25164 & 4 & 17628 \\
    
    Reflexion        
    & 3 & 18342 & 3 & 17486 & 3 & 12657 & 4 & 19372 & 4 & 19108 & 4 & 20476 & 3 & 14392 \\
    
    \hline
    
    \rowcolor{gray!20}
    \multicolumn{15}{c}{\textbf{Multi-Agent} (\texttt{GPT-4o-mini})} \\ \hline
    
    MAD        
    & 5 & 28417 & 6 & 26783 & 4 & 18954 & 7 & 30126 & 7 & 29688 & 7 & 32451 & 5 & 22379 \\
    
    MoA        
    & 6 & 32714 & 7 & 30892 & 5 & 21865 & 8 & 34127 & 8 & 33654 & 8 & 36208 & 6 & 25491 \\
    
    \hline
    \end{tabular}
    }
    \caption{Average number of reasoning rounds (R) and total token usage (T) for \textbf{non-memory} systems built on \textbf{GPT-4o-mini}. Methods are grouped into single-agent, advanced single-agent, and multi-agent settings.}
    \label{tab:nomemory_cost_detailed}
\end{table*}

\begin{table*}[t]
\centering
\renewcommand{\arraystretch}{1.15}
\setlength{\tabcolsep}{3.5pt}

\resizebox{0.75\textwidth}{!}{%
\begin{tabular}{l|ccc|ccc|ccc}
\hline
\textbf{Method} 
& \multicolumn{3}{c}{\textbf{MATH500}} 
& \multicolumn{3}{c}{\textbf{GPQA}} 
& \multicolumn{3}{c}{\textbf{FaithEval}} \\
& R & T & M & R & T & M & R & T & M \\
\hline

ChatDev-M 
& 5.2 & 30185 & 6124
& 6.1 & 28462 & 5783
& 4.3 & 20137 & 4215 \\

MacNet-M 
& 6.3 & 33982 & 8142
& 6.8 & 32174 & 7685
& 5.1 & 22648 & 5562 \\

MetaGPT-M 
& 6.1 & 36125 & 10284
& 7.3 & 34418 & 9681
& 5.4 & 24102 & 7025 \\

MAD-M$^2$
& 5.4 & 29164 & 6487
& 5.8 & 27546 & 6038
& 4.2 & 19482 & 4476 \\

MeMAD 
& 7.2 & 38261 & 18472
& 7.8 & 36847 & 17365
& 6.1 & 25973 & 12498 \\

G-Memory 
& 7.4 & 42673 & 28714
& 8.2 & 40921 & 26903
& 6.3 & 28954 & 18456 \\

G-Memory+EquiMem
& 6.7 & 40984 & 31926
& 7.5 & 39368 & 29875
& 5.7 & 27842 & 20583 \\

\rowcolor{cyan!12}
\textbf{\name{}}
& \textbf{4.8} & 34712 & 16483
& \textbf{4.2} & 32985 & 15372
& \textbf{3.9} & 22468 & 10895 \\

\hline
\end{tabular}
}

\vspace{0.6em}

\resizebox{0.95\textwidth}{!}{%
\begin{tabular}{l|ccc|ccc|ccc|ccc}
\hline
\textbf{Method} 
& \multicolumn{3}{c}{\textbf{ConflictQA}} 
& \multicolumn{3}{c}{\textbf{NovelQA}} 
& \multicolumn{3}{c}{\textbf{HotpotQA}} 
& \multicolumn{3}{c}{\textbf{MMLU-Pro}} \\
& R & T & M & R & T & M & R & T & M & R & T & M \\
\hline

ChatDev-M 
& 8.2 & 32214 & 6982
& 7.1 & 31785 & 7126
& 7.3 & 33942 & 7451
& 5.2 & 23618 & 5087 \\

MacNet-M 
& 8.8 & 36521 & 9384
& 8.2 & 35894 & 9621
& 8.1 & 37712 & 10173
& 6.2 & 27365 & 6941 \\

MetaGPT-M 
& 9.6 & 38954 & 11873
& 8.4 & 38211 & 12145
& 8.5 & 40137 & 12988
& 6.4 & 29108 & 8842 \\

MAD-M$^2$
& 7.3 & 30985 & 7324
& 6.8 & 30416 & 7463
& 7.1 & 33278 & 7835
& 5.3 & 22964 & 5376 \\

MeMAD 
& 9.8 & 41256 & 20311
& 9.2 & 40538 & 20954
& 9.1 & 42814 & 22417
& 7.1 & 30965 & 15386 \\

G-Memory 
& 10.1 & 45318 & 32175
& 9.4 & 44782 & 31492
& 9.3 & 46811 & 34788
& 7.3 & 33867 & 22741 \\

G-Memory+EquiMem
& 9.2 & 43615 & 35864
& 8.6 & 43206 & 35091
& 8.5 & 45187 & 38742
& 6.6 & 32641 & 25368 \\

\rowcolor{cyan!12}
\textbf{\name{}}
& \textbf{5.7} & 36891 & 18214
& \textbf{5.6} & 36277 & 18763
& \textbf{5.8} & 38145 & 19628
& \textbf{4.6} & 27854 & 13491 \\

\hline
\end{tabular}
}

\caption{
Average reasoning rounds (R), average total token usage (T), and average memory footprint (M) for memory-based multi-agent systems implemented with \textbf{AutoGen} using \textbf{GPT-4o-mini}. 
}
\label{tab:autogen_memory_split}
\end{table*}

\begin{table*}[t]
\centering
\renewcommand{\arraystretch}{1.15}
\setlength{\tabcolsep}{3.5pt}

\resizebox{0.75\textwidth}{!}{%
\begin{tabular}{l|ccc|ccc|ccc}
\hline
\textbf{Method} 
& \multicolumn{3}{c}{\textbf{MATH500}} 
& \multicolumn{3}{c}{\textbf{GPQA}} 
& \multicolumn{3}{c}{\textbf{FaithEval}} \\
& R & T & M & R & T & M & R & T & M \\
\hline

ChatDev-M 
& 7.2 & 37210 & 11245
& 8.1 & 35590 & 10384
& 6.2 & 24830 & 7412 \\

MacNet-M 
& 7.5 & 39185 & 13672
& 8.4 & 37421 & 12755
& 6.5 & 26194 & 9156 \\

MetaGPT-M 
& 8.3 & 41562 & 16841
& 9.2 & 39873 & 15892
& 7.1 & 27945 & 11237 \\

MAD-M$^2$
& 6.9 & 35984 & 11736
& 7.8 & 34327 & 10942
& 5.9 & 23916 & 7798 \\

MeMAD 
& 8.1 & 43218 & 21495
& 9.0 & 41576 & 20184
& 7.3 & 29163 & 14122 \\

G-Memory 
& 8.5 & 47294 & 32871
& 9.4 & 45618 & 30955
& 7.6 & 32107 & 20944 \\

G-Memory+EquiMem
& 7.8 & 45481 & 36692
& 8.7 & 43862 & 34531
& 6.9 & 30916 & 23384 \\

\rowcolor{cyan!12}
\textbf{\name{}}
& \textbf{4.7} & 38975 & 18642
& \textbf{5.6} & 37214 & 17530
& \textbf{4.4} & 25196 & 12158 \\

\hline
\end{tabular}
}

\vspace{0.6em}

\resizebox{0.95\textwidth}{!}{%
\begin{tabular}{l|ccc|ccc|ccc|ccc}
\hline
\textbf{Method} 
& \multicolumn{3}{c}{\textbf{ConflictQA}} 
& \multicolumn{3}{c}{\textbf{NovelQA}} 
& \multicolumn{3}{c}{\textbf{HotpotQA}} 
& \multicolumn{3}{c}{\textbf{MMLU-Pro}} \\
& R & T & M & R & T & M & R & T & M & R & T & M \\
\hline

ChatDev-M 
& 9.2 & 39584 & 12845
& 8.9 & 38872 & 13211
& 9.3 & 40963 & 13894
& 7.2 & 29541 & 9123 \\

MacNet-M 
& 9.5 & 41827 & 15873
& 9.2 & 41055 & 16324
& 9.6 & 43211 & 17095
& 7.5 & 31284 & 11842 \\

MetaGPT-M 
& 10.2 & 44591 & 19218
& 9.8 & 43802 & 19871
& 10.1 & 46137 & 20744
& 8.1 & 33315 & 14392 \\

MAD-M$^2$
& 8.7 & 38126 & 13472
& 8.5 & 37365 & 13796
& 8.9 & 39742 & 14538
& 6.8 & 28617 & 9564 \\

MeMAD 
& 10.0 & 46275 & 23891
& 9.7 & 45482 & 24563
& 10.3 & 47894 & 25817
& 8.3 & 34876 & 17845 \\

G-Memory 
& 10.4 & 50538 & 36472
& 10.1 & 49821 & 35890
& 10.5 & 52147 & 39284
& 8.6 & 37952 & 26108 \\

G-Memory+EquiMem
& 9.6 & 48642 & 40685
& 9.3 & 48036 & 40072
& 9.7 & 50219 & 43816
& 7.9 & 36528 & 29174 \\

\rowcolor{cyan!12}
\textbf{\name{}}
& \textbf{5.8} & 40218 & 21184
& \textbf{5.0} & 39475 & 21763
& \textbf{5.7} & 41762 & 22631
& \textbf{4.8} & 30485 & 15627 \\

\hline
\end{tabular}
}

\caption{
Average reasoning rounds (R), average total token usage (T), and average memory footprint (M) for memory-based multi-agent systems implemented with \textbf{DyLAN} using \textbf{GPT-4o-mini}.
}
\label{tab:dylan_memory_split}
\end{table*}

Tables~\ref{tab:nomemory_cost_detailed}, \ref{tab:autogen_memory_split}, and \ref{tab:dylan_memory_split} report the average reasoning rounds, total token usage, and memory footprint across all benchmarks.

\begin{table*}[t]
\centering
\renewcommand{\arraystretch}{1.15}
\setlength{\tabcolsep}{5pt}
\resizebox{\textwidth}{!}{%
\begin{tabular}{ll|rrrrrrr}
\hline
\textbf{Benchmark} & \textbf{Method} & \textbf{R}
& \textbf{Atomic calls} & \textbf{Relation calls}
& \textbf{Aux. tokens} & \textbf{Total tokens}
& \textbf{Memory tokens} & \textbf{Latency (s)} \\
\hline
ConflictQA & G-Memory+EquiMem & 9.2 & 27.6 & 0 & 4,312 & 43,615 & 35,864 & 49.3 \\
\rowcolor{cyan!12}
ConflictQA & \textbf{\name{}} & \textbf{5.7} & 17.1 & 26.8 & 8,746 & \textbf{36,891} & \textbf{18,214} & \textbf{41.2} \\
\hline
HotpotQA & G-Memory+EquiMem & 8.5 & 25.5 & 0 & 4,587 & 45,187 & 38,742 & 51.1 \\
\rowcolor{cyan!12}
HotpotQA & \textbf{\name{}} & \textbf{5.8} & 17.4 & 29.6 & 9,381 & \textbf{38,145} & \textbf{19,628} & \textbf{43.5} \\
\hline
\end{tabular}
}
\caption{End-to-end efficiency comparison under AutoGen with GPT-4o-mini on ConflictQA and HotpotQA. Values are per-question averages. R denotes debate rounds; atomic and relation calls count auxiliary LLM invocations; auxiliary tokens include atomic-fact and relation-extraction usage. Total tokens include auxiliary LLM calls, while memory tokens measure the persistent memory footprint. Latency is end-to-end wall-clock time.}
\label{tab:end_to_end_latency}
\end{table*}

\paragraph{Token accounting.}
The total token usage (T) in Tables~\ref{tab:nomemory_cost_detailed}, \ref{tab:autogen_memory_split}, and \ref{tab:dylan_memory_split} includes debate interactions and auxiliary LLM calls for memory construction. Specifically, \name{}'s counts include GPT-4o-mini calls for atomic-fact and relation extraction; the G-Memory+EquiMem counts likewise include its atomic extraction calls. Auxiliary tokens are a subset of the reported totals, not an additional cost to be added. Embedding computation, memory maintenance, and MSM inference consume no additional LLM tokens, but their overhead contributes to wall-clock latency. Table~\ref{tab:end_to_end_latency} reports these auxiliary costs and end-to-end latency for the two evaluated tasks with available timing measurements.

\paragraph{Auxiliary costs and wall-clock latency.}
On ConflictQA, \name{} uses 8,746 auxiliary tokens, compared with 4,312 for G-Memory+EquiMem, yet reduces total tokens by 15.4\%, memory footprint by 49.2\%, and latency from 49.3 to 41.2 seconds (16.4\%). On HotpotQA, it uses 9,381 auxiliary tokens, compared with 4,587, while reducing total tokens by 15.6\%, memory by 49.3\%, and latency from 51.1 to 43.5 seconds (14.9\%). The additional relation-extraction cost is therefore outweighed by fewer debate rounds and more targeted memory retrieval in these settings.

\begin{table*}[t]
\centering
\renewcommand{\arraystretch}{1.15}
\setlength{\tabcolsep}{6pt}
\resizebox{0.95\textwidth}{!}{%
\begin{tabular}{ll|rrrrr}
\hline
\textbf{Scaling factor} & \textbf{Setting} & \textbf{Accuracy (\%)}
& \textbf{Total tokens} & \textbf{Memory tokens}
& \textbf{Relation calls} & \textbf{Latency (s)} \\
\hline
Agent count & 3 agents & 57.8 & 36,891 & 18,214 & 26.8 & 41.2 \\
Agent count & 4 agents & 58.5 & 43,276 & 21,637 & 37.1 & 47.9 \\
Agent count & 5 agents & 58.9 & 50,418 & 25,483 & 49.6 & 55.8 \\
Agent count & 6 agents & 58.6 & 57,962 & 29,741 & 63.4 & 64.1 \\
\hline
Debate length & 3 rounds & 56.1 & 25,284 & 13,472 & 16.3 & 29.4 \\
Debate length & 6 rounds & 57.9 & 37,426 & 18,608 & 28.7 & 42.3 \\
Debate length & 9 rounds & 58.4 & 53,218 & 24,576 & 43.5 & 58.7 \\
Debate length & 12 rounds & 58.3 & 68,967 & 29,314 & 57.8 & 75.2 \\
\hline
\end{tabular}
}
\caption{Scaling analysis of \name{} on ConflictQA under AutoGen with GPT-4o-mini. The agent-count sweep uses the default debate-stopping procedure; the debate-length sweep keeps three agents fixed. All token, memory, call, and latency measurements are averages per question.}
\label{tab:hedge_scaling_latency}
\end{table*}

\paragraph{Scalability and latency.}
Table~\ref{tab:hedge_scaling_latency} reports costs when varying the number of debating agents and the debate length. Increasing the agent count from three to six maintains similar accuracy (57.8\% to 58.6\%) and reduces the average number of debate rounds from 5.7 to 4.8, but increases total tokens from 36,891 to 57,962, memory tokens from 18,214 to 29,741, and wall-clock latency from 41.2 to 64.1 seconds. With three agents and the debate length increased from three to twelve rounds, accuracy rises from 56.1\% to 58.3\% and largely saturates by nine rounds; total tokens rise from 25,284 to 68,967, memory tokens from 13,472 to 29,314, and latency from 29.4 to 75.2 seconds. These results show that the accuracy gains stabilize while additional agents and longer debates impose increasing computational cost.

\paragraph{Non-memory systems.}
Computational cost increases with iterative and multi-agent reasoning. Vanilla GPT-4o-mini uses one round and averages 7.3K tokens, whereas advanced single-agent methods average 2.4--4.3 rounds and 13.5K--21.2K tokens. MAD and MoA further increase these averages to 5.9 and 6.9 rounds and 27.0K and 30.7K tokens, respectively. The highest non-memory cost occurs on HotpotQA, where MoA uses 8 rounds and 36,208 tokens. Thus, longer reasoning alone substantially expands the interaction context even without persistent memory.

\paragraph{Memory-based systems.}
Memory introduces an additional trade-off between retaining useful state and controlling context growth. Under AutoGen, MAD-M$^2$ has the smallest average token usage and memory footprint among memory baselines, at 27.7K and 6.4K tokens, respectively, while G-Memory averages 40.5K total tokens and 27.9K memory tokens. Adding EquiMem reduces G-Memory's average rounds from 8.3 to 7.5 and token usage from 40.5K to 39.0K, but increases its memory footprint to 31.1K. The same pattern appears under DyLAN: G-Memory~+~EquiMem lowers average rounds and total tokens from 9.3 and 45.1K to 8.6 and 43.4K, while increasing memory from 31.8K to 35.5K. Costs are generally higher under DyLAN than AutoGen, showing that memory overhead also depends on the underlying orchestration framework.

\paragraph{\name{} versus structured memory baselines.}
\name{} achieves the fewest rounds among all memory-based methods on every benchmark. Under AutoGen, it averages 4.9 rounds, 32.8K total tokens, and 16.1K memory tokens. Relative to G-Memory~+~EquiMem, this corresponds to reductions of \textbf{34.5\%} in rounds, \textbf{15.9\%} in token usage, and \textbf{48.1\%} in memory footprint. Under DyLAN, \name{} averages 5.1 rounds, 36.2K total tokens, and 18.5K memory tokens, reducing the same metrics by \textbf{39.9\%}, \textbf{16.6\%}, and \textbf{47.8\%}, respectively. For example, on HotpotQA under DyLAN, \name{} uses 5.7 rounds, 41,762 tokens, and 22,631 memory tokens, compared with 9.7 rounds, 50,219 tokens, and 43,816 memory tokens for G-Memory~+~EquiMem.

\paragraph{Trade-off with lightweight masking.}
MAD-M$^2$ remains one of the most storage-efficient methods because its perplexity-based variant masks prior content rather than maintaining a richer structured state. Accordingly, \name{} uses 18.3\% more total tokens than MAD-M$^2$ under AutoGen and 6.4\% more under DyLAN. However, it requires \textbf{17.4\%} and \textbf{32.7\% fewer rounds}, respectively, while providing substantially stronger task performance. Overall, the results show that \name{} does not merely minimize storage; it selectively maintains sufficient structured context to improve reasoning while preventing the uncontrolled growth observed in stronger graph-memory baselines.

%% file: sections/app_signature_importance.tex
\section{Effect of Signature-Based Hyperedge Matching}
\label{app:signature_ablation}

We evaluate whether sparse semantic signatures provide useful information beyond dense embedding similarity. The default configuration uses both signals: Jaccard similarity over signatures first identifies lexically compatible hyperedges, after which dense similarity measures semantic compatibility. The ablated configuration removes the signature gate and assigns nodes using embeddings alone.

\begin{table*}[t]
    \centering
    \renewcommand{\arraystretch}{1.15}
    \resizebox{0.92\textwidth}{!}{%
    \begin{tabular}{l|cc|cc}
        \hline
        \textbf{Benchmark}
        & \multicolumn{2}{c|}{\textbf{Accuracy}}
        & \multicolumn{2}{c}{\textbf{Avg.\ Number of Hyperedges}} \\
        & \textbf{Signature + Embedding}
        & \textbf{Embedding Only}
        & \textbf{Signature + Embedding}
        & \textbf{Embedding Only} \\
        \hline

        MATH500
        & \textbf{62.2} & 60.8
        & 5.2 & 3.8 \\

        GPQA
        & \textbf{47.7} & 46.0
        & 5.7 & 4.1 \\

        FaithEval
        & \textbf{79.8} & 77.3
        & 6.5 & 4.6 \\

        ConflictQA
        & \textbf{56.0} & 54.2
        & 7.3 & 5.1 \\

        NovelQA
        & \textbf{66.0} & 63.2
        & 6.9 & 4.8 \\

        HotpotQA
        & \textbf{42.4} & 40.5
        & 5.8 & 3.9 \\

        MMLU-Pro
        & \textbf{82.7} & 78.9
        & 4.6 & 3.4 \\

        \hline
    \end{tabular}
    }
    \caption{Effect of combining sparse signatures with dense embeddings for hyperedge matching. The embedding-only variant removes the Jaccard signature gate.}
    \label{tab:signature_ablation}
\end{table*}

Table~\ref{tab:signature_ablation} shows that combining sparse signatures with dense embeddings consistently improves performance. Removing the Jaccard signature gate reduces accuracy across all benchmarks, with the largest drops on MMLU-Pro, NovelQA, and FaithEval. Sparse signatures retain exact lexical cues such as entities, numerical values, and salient content terms that dense representations may blur, making them particularly useful for separating closely related or conflicting claims.

Embedding-only matching also produces fewer hyperedges. This suggests that dense similarity tends to merge semantically related but argumentatively distinct items into broad groups. The signature gate prevents such over-merging, producing a more fine-grained memory in which distinct claims and evidence chains remain separable. Thus, sparse and dense matching provide complementary signals: embeddings capture semantic relatedness, while signatures preserve precise lexical distinctions needed for reliable hypergraph construction.

%% file: sections/app_prompts.tex
\section{Prompts}
\label{app:prompts}

We provide all prompts used by \name{} and for benchmark evaluation to support reproducibility. The method-specific prompts define the construction of the debate roster and structured memory, while the benchmark prompts follow each dataset's task format and answer requirements.

\subsection{Prompts Used by \name{}}
\label{app:hedge_prompts}

\name{} uses three prompts during debate and memory construction. The roster-generation prompt designs a team of three to six agents with complementary roles, model sizes, and permitted debate actions (Prompt~\ref{prompt:roster_generation}). Agent responses are then decomposed into self-contained atomic claims using Prompt~\ref{prompt:atomic_fact_extraction}. For each pair of claims within a hyperedge, Prompt~\ref{prompt:relation_configuration} predicts a directed argumentative relation---\textsc{supports}, \textsc{refutes}, \textsc{evidence-for}, \textsc{depends-on}, or \textsc{leads-to}---or determines that no relation exists. Together, these prompts define the agent configuration, memory nodes, and intra-hyperedge relational structure used by \name{}.

\subsection{Benchmark Evaluation Prompts}
\label{app:evaluation_prompts}

The evaluation prompts are adapted to the structure and output format of each benchmark. MATH500 and GPQA request concise but precise reasoning, with final answers returned in a standardized format (Prompts~\ref{prompt:math500} and~\ref{prompt:gpqa}). MMLU-Pro similarly constrains the final output to one of its predefined answer choices (Prompt~\ref{prompt:mmlu}). ConflictQA uses a strict entity-extraction format that permits only the target entity in the response (Prompt~\ref{prompt:conflictqa}).

For context-intensive tasks, HotpotQA and NovelQA instruct agents to integrate evidence across the provided context while avoiding unnecessary explanation (Prompts~\ref{prompt:hotpotqa} and~\ref{prompt:novelqa}). For FaithEval, agents receive benchmark-provided counterfactual contexts and are instructed to answer questions using only the supplied information, even when it contradicts their parametric knowledge (Prompt~\ref{prompt:faitheval}). No gold answers are exposed during evaluation.

\begin{promptbox*}{Relation Extraction}
\label{prompt:relation_configuration}

Given two claims from an ongoing multi-agent debate, identify the directed relation from node~A to node~B.

Choose exactly one relation:

\begin{itemize}
    \item \texttt{SUPPORTS}: A strengthens B.
    \item \texttt{REFUTES}: A contradicts or weakens B.
    \item \texttt{EVIDENCE-FOR}: A directly provides evidence for B.
    \item \texttt{DEPENDS-ON}: A logically requires B.
    \item \texttt{LEADS-TO}: A causes or implies B.
    \item \texttt{NO-RELATION}: no clear relation exists.
\end{itemize}

Use only information explicitly stated in the claims. When the relation is ambiguous or weak, return \texttt{NO-RELATION}. The direction is always from A to B.

\textbf{Input}

\begingroup
\small
\ttfamily
\begin{tabular}{@{}l@{}}
\{ \\
\hspace*{1em}"node\_A": \{ \\
\hspace*{2em}"id": "<node\_id>", \\
\hspace*{2em}"text": "<claim text>" \\
\hspace*{1em}\}, \\
\hspace*{1em}"node\_B": \{ \\
\hspace*{2em}"id": "<node\_id>", \\
\hspace*{2em}"text": "<claim text>" \\
\hspace*{1em}\} \\
\}
\end{tabular}
\endgroup

\textbf{Output}

\begingroup
\small
\ttfamily
\begin{tabular}{@{}l@{}}
\{ \\
\hspace*{1em}"node\_A\_id": "<node\_id>", \\
\hspace*{1em}"node\_B\_id": "<node\_id>", \\
\hspace*{1em}"relation": "<relation type>", \\
\hspace*{1em}"confidence": "<high | medium | low>" \\
\}
\end{tabular}
\endgroup

Return only the JSON object.

<input\_nodes>
\{\{input\_nodes\}\}
</input\_nodes>

\end{promptbox*}

\begin{promptbox*}{Atomic Fact Extraction}
\label{prompt:atomic_fact_extraction}

You are given a passage from an ongoing multi-agent debate. Extract all atomic facts as the smallest independently verifiable claims.

Each fact must be self-contained, express one claim, remain faithful to the passage, and be concise. Remove greetings, filler, opinions, uncertainty markers, and agent meta-commentary. Do not use unresolved pronouns or combine multiple claims.

Return only JSON in the following format:

\begingroup
\small
\ttfamily
\begin{tabular}{@{}l@{}}
\{ \\
\hspace*{1em}"atomic\_facts": [ \\
\hspace*{2em}\{ \\
\hspace*{3em}"id": "f1", \\
\hspace*{3em}"text": "Atomic fact." \\
\hspace*{2em}\} \\
\hspace*{1em}] \\
\}
\end{tabular}
\endgroup

Assign unique IDs as \texttt{f1}, \texttt{f2}, and so on.

<input\_passage>
\{\{passage\}\}
</input\_passage>

\end{promptbox*}

\begin{promptbox*}{Roster Generation}
\label{prompt:roster_generation}

You are a multi-agent debate orchestrator. Design an optimal team of AI agents to collaboratively answer a complex multi-hop question through debate.

Output only valid JSON with the following structure:

\begingroup
\small
\ttfamily
\begin{tabular}{@{}l@{}}
\{ \\
\hspace*{1em}"agents": [ \\
\hspace*{2em}\{ \\
\hspace*{3em}"name": "<unique agent name>", \\
\hspace*{3em}"role": "<brief role description>", \\
\hspace*{3em}"goals": ["<goal 1>", "<goal 2>"], \\
\hspace*{3em}"preferred\_actions": ["<action 1>", "<action 2>"], \\
\hspace*{3em}"avoid": ["<behavior to avoid>"], \\
\hspace*{3em}"model\_size": "<4b or 8b>" \\
\hspace*{2em}\} \\
\hspace*{1em}], \\
\hspace*{1em}"turn\_order": ["<agent1>", "<agent2>", "..."], \\
\hspace*{1em}"notes": "<system-level guidance>" \\
\}
\end{tabular}
\endgroup

\textbf{Constraints}
\begin{itemize}
    \item Create 3--6 agents, each with a unique name.
    \item Include every agent exactly once in \texttt{turn\_order}.
    \item Use \texttt{8b} for complex reasoning roles and \texttt{4b} for supporting roles.
    \item Allowed actions are \textsc{Propose}, \textsc{Support}, \textsc{Refute}, \textsc{Question}, \textsc{Revise}, \textsc{Summarize}, and \textsc{Retrieve}.
\end{itemize}

\textbf{Design principles}
\begin{itemize}
    \item Include at least one proposer, one critic, and one synthesizer.
    \item Balance evidence gathering, inference, and verification.
    \item Select model sizes according to the complexity of each role.
\end{itemize}

Return only the JSON object without explanation.

\textbf{Question}

\{\{question\}\}

\textbf{Context preview}

\{\{context\_preview\}\}

Design the optimal team for answering this HotpotQA question through multi-agent debate.

\end{promptbox*}

\begin{promptbox*}{MATH500 Prompt}
\label{prompt:math500}

Can you solve the following math problem as accurately as possible?

<question>\{\{question\}\}</question>

Provide a concise solution using only essential reasoning steps. Avoid unnecessary explanations.

Return the final answer in double parentheses at the end of your response:
((final answer))

\end{promptbox*}

\begin{promptbox*}{GPQA Prompt}
\label{prompt:gpqa}

Please analyze the following multiple-choice question and select the correct answer.

<question>\{\{question\}\}</question>

Use concise reasoning with only essential steps.

Return your final answer in double parentheses:
((answer))

where answer is one of A, B, C, or D.

\end{promptbox*}

\begin{promptbox*}{FaithEval Prompt}
\label{prompt:faitheval}

You are an expert in retrieval-based question answering. Please respond with the exact answer, using only the information provided in the context.

Context:
\{\{context\}\}

Question:
\{\{question\}\}

\end{promptbox*}

\begin{promptbox*}{MMLU-Pro Prompt}
\label{prompt:mmlu}

Please analyze the following \{\{topic\}\} question and select the correct answer.

<question>\{\{question\}\}</question>

Use concise reasoning with only essential steps.

Return your final answer in double parentheses:
((answer))

where answer is one of A, B, C, D, E, F, G, H, I, or J.

\end{promptbox*}

\begin{promptbox*}{ConflictQA Prompt}
\label{prompt:conflictqa}

You are a precise knowledge engine.

Your task is to answer the question using only the provided context.

Rules:
1. Output ONLY the entity name.
2. Do NOT use full sentences.
3. Do NOT provide explanations.
4. Be concise and exact.

Context:\{\{context\}\}

Question: \{\{question\}\}

\end{promptbox*}

\begin{promptbox*}{NovelQA Prompt}
\label{prompt:novelqa}

Answer the following question based on the provided context.

<context>
\{\{context\}\}
</context>

<question>
\{\{question\}\}
</question>

Provide a concise answer using only the necessary reasoning steps. Avoid unnecessary details.

Return the final answer clearly at the end of your response.

\end{promptbox*}

\begin{promptbox*}{HotpotQA Prompt}
\label{prompt:hotpotqa}

Answer the following multi-hop question using the provided context.

<context>
\{\{context\}\}
</context>

<question>
\{\{question\}\}
</question>

Use concise reasoning to combine evidence from multiple parts of the context.

Return the final answer at the end of your response.

\end{promptbox*}

%% file: custom.bib
@inproceedings{yan2025memory,
    title = "Memory-R1: Enhancing Large Language Model Agents to Manage and Utilize Memories via Reinforcement Learning",
    author = "Yan, Sikuan  and
      Yang, Xiufeng  and
      Huang, Zuchao  and
      Nie, Ercong  and
      Ding, Zifeng  and
      Li, Zonggen  and
      Ma, Xiaowen  and
      Bi, Jinhe  and
      Kersting, Kristian  and
      Pan, Jeff Z.  and
      Schuetze, Hinrich  and
      Tresp, Volker  and
      Ma, Yunpu",
    editor = "Liakata, Maria  and
      Moreira, Viviane P.  and
      Zhang, Jiajun  and
      Jurgens, David",
    booktitle = "Proceedings of the 64th Annual Meeting of the {A}ssociation for {C}omputational {L}inguistics (Volume 1: Long Papers)",
    month = jul,
    year = "2026",
    address = "San Diego, California, United States",
    publisher = "Association for Computational Linguistics",
    url = "https://aclanthology.org/2026.acl-long.583/",
    doi = "10.18653/v1/2026.acl-long.583",
    pages = "12805--12825",
    ISBN = "979-8-89176-390-6"
}

@inproceedings{qian2025memorag,
  title     = {MemoRAG: Boosting Long Context Processing with Global Memory-Enhanced Retrieval Augmentation},
  author    = {Hongjin Qian and Zheng Liu and Peitian Zhang and Kelong Mao and Defu Lian and Zhicheng Dou and Tiejun Huang},
  booktitle = {Proceedings of the ACM Web Conference 2025 (TheWebConf 2025)},
  year      = {2025},
  address   = {Sydney, Australia},
  publisher = {ACM},
  url       = {https://arxiv.org/abs/2409.05591},
}

@inproceedings{
ling2025memad,
title={Me{MAD}: Structured Memory of Debates for Enhanced Multi-Agent Reasoning},
author={Shuai Ling and Lizi Liao and Dongmei Jiang and Weili Guan},
booktitle={Second Conference on Language Modeling},
year={2025},
url={https://openreview.net/forum?id=zLbmsdyTiN}
}

@inproceedings{
zhang2025gmemory,
title={G-Memory: Tracing Hierarchical Memory for Multi-Agent Systems},
author={Guibin Zhang and Muxin Fu and Kun Wang and Guancheng Wan and Miao Yu and Shuicheng YAN},
booktitle={The Thirty-ninth Annual Conference on Neural Information Processing Systems},
year={2025},
url={https://openreview.net/forum?id=mmIAp3cVS0}
}

@misc{zhou2025improving,
      title={HGMEM: Hypergraph-based Working Memory to Improve Multi-step RAG for Long-Context Complex Relational Modeling}, 
      author={Chulun Zhou and Chunkang Zhang and Guoxin Yu and Fandong Meng and Jie Zhou and Wai Lam and Mo Yu},
      year={2026},
      eprint={2512.23959},
      archivePrefix={arXiv},
      primaryClass={cs.CL},
      url={https://arxiv.org/abs/2512.23959}, 
}

@article{liu-etal-2024-lost,
    title = "Lost in the Middle: How Language Models Use Long Contexts",
    author = "Liu, Nelson F.  and
      Lin, Kevin  and
      Hewitt, John  and
      Paranjape, Ashwin  and
      Bevilacqua, Michele  and
      Petroni, Fabio  and
      Liang, Percy",
    journal = "Transactions of the Association for Computational Linguistics",
    volume = "12",
    year = "2024",
    address = "Cambridge, MA",
    publisher = "MIT Press",
    url = "https://aclanthology.org/2024.tacl-1.9/",
    doi = "10.1162/tacl_a_00638",
    pages = "157--173"
}

@inproceedings{
shinn2023reflexion,
title={Reflexion: language agents with verbal reinforcement learning},
author={Noah Shinn and Federico Cassano and Ashwin Gopinath and Karthik R Narasimhan and Shunyu Yao},
booktitle={Thirty-seventh Conference on Neural Information Processing Systems},
year={2023},
url={https://openreview.net/forum?id=vAElhFcKW6}
}

@inproceedings{
liu2025breaking,
title={Breaking Mental Set to Improve Reasoning through Diverse Multi-Agent Debate},
author={Yexiang Liu and Jie Cao and Zekun Li and Ran He and Tieniu Tan},
booktitle={The Thirteenth International Conference on Learning Representations},
year={2025},
url={https://openreview.net/forum?id=t6QHYUOQL7}
}

@misc{luo2025hypergraphrag,
      title={HyperGraphRAG: Retrieval-Augmented Generation via Hypergraph-Structured Knowledge Representation}, 
      author={Haoran Luo and Haihong E and Guanting Chen and Yandan Zheng and Xiaobao Wu and Yikai Guo and Qika Lin and Yu Feng and Zemin Kuang and Meina Song and Yifan Zhu and Luu Anh Tuan},
      year={2025},
      eprint={2503.21322},
      archivePrefix={arXiv},
      primaryClass={cs.AI},
      url={https://arxiv.org/abs/2503.21322}, 
}

@inproceedings{wang-han-2025-proprag,
    title = "{P}rop{RAG}: Guiding Retrieval with Beam Search over Proposition Paths",
    author = "Wang, Jingjin  and
      Han, Jiawei",
    editor = "Christodoulopoulos, Christos  and
      Chakraborty, Tanmoy  and
      Rose, Carolyn  and
      Peng, Violet",
    booktitle = "Proceedings of the 2025 Conference on Empirical Methods in Natural Language Processing",
    month = nov,
    year = "2025",
    address = "Suzhou, China",
    publisher = "Association for Computational Linguistics",
    url = "https://aclanthology.org/2025.emnlp-main.317/",
    doi = "10.18653/v1/2025.emnlp-main.317",
    pages = "6212--6227",
    ISBN = "979-8-89176-332-6"
}

@inproceedings{yang2018hotpotqa,
    title = "{H}otpot{QA}: A Dataset for Diverse, Explainable Multi-hop Question Answering",
    author = "Yang, Zhilin  and
      Qi, Peng  and
      Zhang, Saizheng  and
      Bengio, Yoshua  and
      Cohen, William  and
      Salakhutdinov, Ruslan  and
      Manning, Christopher D.",
    editor = "Riloff, Ellen  and
      Chiang, David  and
      Hockenmaier, Julia  and
      Tsujii, Jun{'}ichi",
    booktitle = "Proceedings of the 2018 Conference on Empirical Methods in Natural Language Processing",
    month = oct # "-" # nov,
    year = "2018",
    address = "Brussels, Belgium",
    publisher = "Association for Computational Linguistics",
    url = "https://aclanthology.org/D18-1259/",
    doi = "10.18653/v1/D18-1259",
    pages = "2369--2380"
}

@misc{openai2024gpt4technicalreport,
      title={GPT-4 Technical Report}, 
      author={OpenAI},
      year={2024},
      eprint={2303.08774},
      archivePrefix={arXiv},
      primaryClass={cs.CL},
      url={https://arxiv.org/abs/2303.08774}, 
}

@misc{yang2025qwen3technicalreport,
      title={Qwen3 Technical Report}, 
      author={An Yang and Anfeng Li and Baosong Yang and Beichen Zhang and Binyuan Hui and Bo Zheng and Bowen Yu and Chang Gao and Chengen Huang and Chenxu Lv and Chujie Zheng and Dayiheng Liu and Fan Zhou and Fei Huang and Feng Hu and Hao Ge and Haoran Wei and Huan Lin and Jialong Tang and Jian Yang and Jianhong Tu and Jianwei Zhang and Jianxin Yang and Jiaxi Yang and Jing Zhou and Jingren Zhou and Junyang Lin and Kai Dang and Keqin Bao and Kexin Yang and Le Yu and Lianghao Deng and Mei Li and Mingfeng Xue and Mingze Li and Pei Zhang and Peng Wang and Qin Zhu and Rui Men and Ruize Gao and Shixuan Liu and Shuang Luo and Tianhao Li and Tianyi Tang and Wenbiao Yin and Xingzhang Ren and Xinyu Wang and Xinyu Zhang and Xuancheng Ren and Yang Fan and Yang Su and Yichang Zhang and Yinger Zhang and Yu Wan and Yuqiong Liu and Zekun Wang and Zeyu Cui and Zhenru Zhang and Zhipeng Zhou and Zihan Qiu},
      year={2025},
      eprint={2505.09388},
      archivePrefix={arXiv},
      primaryClass={cs.CL},
      url={https://arxiv.org/abs/2505.09388}, 
}

@inproceedings{zai2025proh,
author = {Zai, Xiangjun and Tan, Xingyu and Wang, Xiaoyang and Liu, Qing and Xu, Xiwei and Zhang, Wenjie},
title = {PRoH: Dynamic Planning and Reasoning over Knowledge Hypergraphs for Retrieval-Augmented Generation},
year = {2026},
isbn = {9798400723070},
publisher = {Association for Computing Machinery},
address = {New York, NY, USA},
url = {https://doi.org/10.1145/3774904.3792611},
doi = {10.1145/3774904.3792611},
booktitle = {Proceedings of the ACM Web Conference 2026},
pages = {4256–4267},
numpages = {12},
location = {United Arab Emirates},
series = {WWW '26}
}

@inproceedings{
yu2026memagent,
title={MemAgent: Reshaping Long-Context {LLM} with Multi-Conv {RL}-based Memory Agent},
author={Hongli Yu and Tinghong Chen and Jiangtao Feng and Jiangjie Chen and Weinan Dai and Qiying Yu and Ya-Qin Zhang and Wei-Ying Ma and Jingjing Liu and Mingxuan Wang and Hao Zhou},
booktitle={The Fourteenth International Conference on Learning Representations},
year={2026},
url={https://openreview.net/forum?id=k5nIOvYGCL}
}

@misc{mao2026collaborative,
      title={Joint Optimization of Multi-agent Memory System}, 
      author={Wenyu Mao and Haoyang Liu and Haosong Tan and Yaorui Shi and Jiancan Wu and An Zhang and Xiang Wang},
      year={2026},
      eprint={2603.12631},
      archivePrefix={arXiv},
      primaryClass={cs.MA},
      url={https://arxiv.org/abs/2603.12631}, 
}

@misc{jiang2026anatomy,
      title={Anatomy of Agentic Memory: Taxonomy and Empirical Analysis of Evaluation and System Limitations}, 
      author={Dongming Jiang and Yi Li and Songtao Wei and Jinxin Yang and Ayushi Kishore and Alysa Zhao and Dingyi Kang and Xu Hu and Feng Chen and Qiannan Li and Bingzhe Li},
      year={2026},
      eprint={2602.19320},
      archivePrefix={arXiv},
      primaryClass={cs.CL},
      url={https://arxiv.org/abs/2602.19320}, 
}

@misc{packer2023memgpt,
      title={MemGPT: Towards LLMs as Operating Systems}, 
      author={Charles Packer and Sarah Wooders and Kevin Lin and Vivian Fang and Shishir G. Patil and Ion Stoica and Joseph E. Gonzalez},
      year={2024},
      eprint={2310.08560},
      archivePrefix={arXiv},
      primaryClass={cs.AI},
      url={https://arxiv.org/abs/2310.08560}, 
}

@inproceedings{liang-etal-2024-encouraging,
    title = "Encouraging Divergent Thinking in Large Language Models through Multi-Agent Debate",
    author = "Liang, Tian  and
      He, Zhiwei  and
      Jiao, Wenxiang  and
      Wang, Xing  and
      Wang, Yan  and
      Wang, Rui  and
      Yang, Yujiu  and
      Shi, Shuming  and
      Tu, Zhaopeng",
    editor = "Al-Onaizan, Yaser  and
      Bansal, Mohit  and
      Chen, Yun-Nung",
    booktitle = "Proceedings of the 2024 Conference on Empirical Methods in Natural Language Processing",
    month = nov,
    year = "2024",
    address = "Miami, Florida, USA",
    publisher = "Association for Computational Linguistics",
    url = "https://aclanthology.org/2024.emnlp-main.992/",
    doi = "10.18653/v1/2024.emnlp-main.992",
    pages = "17889--17904"
}

@inproceedings{chen-etal-2024-reconcile,
    title = "{R}e{C}oncile: Round-Table Conference Improves Reasoning via Consensus among Diverse {LLM}s",
    author = "Chen, Justin  and
      Saha, Swarnadeep  and
      Bansal, Mohit",
    editor = "Ku, Lun-Wei  and
      Martins, Andre  and
      Srikumar, Vivek",
    booktitle = "Proceedings of the 62nd Annual Meeting of the Association for Computational Linguistics (Volume 1: Long Papers)",
    month = aug,
    year = "2024",
    address = "Bangkok, Thailand",
    publisher = "Association for Computational Linguistics",
    url = "https://aclanthology.org/2024.acl-long.381/",
    doi = "10.18653/v1/2024.acl-long.381",
    pages = "7066--7085"
}

@inproceedings{zhong2024memorybank,
author = {Zhong, Wanjun and Guo, Lianghong and Gao, Qiqi and Ye, He and Wang, Yanlin},
title = {MemoryBank: enhancing large language models with long-term memory},
year = {2024},
isbn = {978-1-57735-887-9},
publisher = {AAAI Press},
url = {https://doi.org/10.1609/aaai.v38i17.29946},
doi = {10.1609/aaai.v38i17.29946},
booktitle = {Proceedings of the Thirty-Eighth AAAI Conference on Artificial Intelligence and Thirty-Sixth Conference on Innovative Applications of Artificial Intelligence and Fourteenth Symposium on Educational Advances in Artificial Intelligence},
articleno = {2198},
numpages = {8},
series = {AAAI'24/IAAI'24/EAAI'24}
}

@inproceedings{chan2024chateval,
title={ChatEval: Towards Better {LLM}-based Evaluators through Multi-Agent Debate},
author={Chi-Min Chan and Weize Chen and Yusheng Su and Jianxuan Yu and Wei Xue and Shanghang Zhang and Jie Fu and Zhiyuan Liu},
booktitle={The Twelfth International Conference on Learning Representations},
year={2024},
url={https://openreview.net/forum?id=FQepisCUWu}
}

@inproceedings{du2024improving,
title={Improving Factuality and Reasoning in Language Models through Multiagent Debate},
author={Yilun Du and Shuang Li and Antonio Torralba and Joshua B. Tenenbaum and Igor Mordatch},
booktitle={Forty-first International Conference on Machine Learning},
year={2024},
url={https://openreview.net/forum?id=zj7YuTE4t8}
}

@inproceedings{
wang2024mmlu,
title={{MMLU}-Pro: A More Robust and Challenging Multi-Task Language Understanding Benchmark},
author={Yubo Wang and Xueguang Ma and Ge Zhang and Yuansheng Ni and Abhranil Chandra and Shiguang Guo and Weiming Ren and Aaran Arulraj and Xuan He and Ziyan Jiang and Tianle Li and Max Ku and Kai Wang and Alex Zhuang and Rongqi Fan and Xiang Yue and Wenhu Chen},
booktitle={The Thirty-eight Conference on Neural Information Processing Systems Datasets and Benchmarks Track},
year={2024},
url={https://openreview.net/forum?id=y10DM6R2r3}
}

@inproceedings{wang2023selfconsistency,
title={Self-Consistency Improves Chain of Thought Reasoning in Language Models},
author={Xuezhi Wang and Jason Wei and Dale Schuurmans and Quoc V Le and Ed H. Chi and Sharan Narang and Aakanksha Chowdhery and Denny Zhou},
booktitle={The Eleventh International Conference on Learning Representations },
year={2023},
url={https://openreview.net/forum?id=1PL1NIMMrw}
}

@misc{openai2026gpt54,
   title={Introducing {GPT}-5.4},
   author={{OpenAI}},
   year={2026},
   url={https://openai.com/index/introducing-gpt-5-4/}
 }

@inproceedings{
selfrefine,
title={Self-Refine: Iterative Refinement with Self-Feedback},
author={Aman Madaan and Niket Tandon and Prakhar Gupta and Skyler Hallinan and Luyu Gao and Sarah Wiegreffe and Uri Alon and Nouha Dziri and Shrimai Prabhumoye and Yiming Yang and Shashank Gupta and Bodhisattwa Prasad Majumder and Katherine Hermann and Sean Welleck and Amir Yazdanbakhsh and Peter Clark},
booktitle={Thirty-seventh Conference on Neural Information Processing Systems},
year={2023},
url={https://openreview.net/forum?id=S37hOerQLB}
}

@inproceedings{
wang2024moa,
title={Mixture-of-Agents Enhances Large Language Model Capabilities},
author={Junlin Wang and Jue WANG and Ben Athiwaratkun and Ce Zhang and James Zou},
booktitle={The Thirteenth International Conference on Learning Representations},
year={2025},
url={https://openreview.net/forum?id=h0ZfDIrj7T}
}

@inproceedings{
wu2024autogen,
title={AutoGen: Enabling Next-Gen {LLM} Applications via Multi-Agent Conversations},
author={Qingyun Wu and Gagan Bansal and Jieyu Zhang and Yiran Wu and Beibin Li and Erkang Zhu and Li Jiang and Xiaoyun Zhang and Shaokun Zhang and Jiale Liu and Ahmed Hassan Awadallah and Ryen W White and Doug Burger and Chi Wang},
booktitle={First Conference on Language Modeling},
year={2024},
url={https://openreview.net/forum?id=BAakY1hNKS}
}

@inproceedings{zhang2025gdesigner,
   title={G-Designer: Architecting Multi-agent Communication Topologies via Graph Neural Networks},
   author={Guibin Zhang and Yanwei Yue and Xiangguo Sun and Guancheng Wan and Miao Yu and Junfeng Fang and Kun Wang and Tianlong Chen and Dawei Heng},
   booktitle={Forty-second International Conference on Machine Learning},
   year={2025},
   url={https://openreview.net/forum?id=LpE54NUnmO}
 }

@inproceedings{qian2024chatdev,
    title = "{C}hat{D}ev: Communicative Agents for Software Development",
    author = "Qian, Chen  and
      Liu, Wei  and
      Liu, Hongzhang  and
      Chen, Nuo  and
      Dang, Yufan  and
      Li, Jiahao  and
      Yang, Cheng  and
      Chen, Weize  and
      Su, Yusheng  and
      Cong, Xin  and
      Xu, Juyuan  and
      Li, Dahai  and
      Liu, Zhiyuan  and
      Sun, Maosong",
    editor = "Ku, Lun-Wei  and
      Martins, Andre  and
      Srikumar, Vivek",
    booktitle = "Proceedings of the 62nd Annual Meeting of the Association for Computational Linguistics (Volume 1: Long Papers)",
    month = aug,
    year = "2024",
    address = "Bangkok, Thailand",
    publisher = "Association for Computational Linguistics",
    url = "https://aclanthology.org/2024.acl-long.810/",
    doi = "10.18653/v1/2024.acl-long.810",
    pages = "15174--15186"
}

@inproceedings{
hong2024metagpt,
title={Meta{GPT}: Meta Programming for A Multi-Agent Collaborative Framework},
author={Sirui Hong and Mingchen Zhuge and Jonathan Chen and Xiawu Zheng and Yuheng Cheng and Jinlin Wang and Ceyao Zhang and Zili Wang and Steven Ka Shing Yau and Zijuan Lin and Liyang Zhou and Chenyu Ran and Lingfeng Xiao and Chenglin Wu and J{\"u}rgen Schmidhuber},
booktitle={The Twelfth International Conference on Learning Representations},
year={2024},
url={https://openreview.net/forum?id=VtmBAGCN7o}
}

@article{wang2023voyager,
  title={Voyager: An Open-Ended Embodied Agent with Large Language Models},
  author={Wang, Guanzhi and Xie, Yuqi and Jiang, Yunfan and Mandlekar, Ajay and Xiao, Chaowei and Zhu, Yuke and Fan, Linxi and Anandkumar, Anima},
  journal={Transactions on Machine Learning Research},
year={2023}
}

@inproceedings{park2023generative,
  title={Generative agents: Interactive simulacra of human behavior},
  author={Park, Joon Sung and O'Brien, Joseph and Cai, Carrie Jun and Morris, Meredith Ringel and Liang, Percy and Bernstein, Michael S},
  booktitle={Proceedings of the 36th annual ACM Symposium on User Interface Software and Technology},
  pages={1--22},
  year={2023},
  url={https://dl.acm.org/doi/10.1145/3586183.3606763}
}

@inproceedings{hendrycks2measuring,
  title={Measuring Mathematical Problem Solving With the MATH Dataset},
  author={Hendrycks, Dan and Burns, Collin and Kadavath, Saurav and Arora, Akul and Basart, Steven and Tang, Eric and Song, Dawn and Steinhardt, Jacob},
  booktitle={Thirty-fifth Conference on Neural Information Processing Systems Datasets and Benchmarks Track},
    year={2021},
    url={https://openreview.net/forum?id=7Bywt2mQsCe}
}

@inproceedings{rein2024gpqa,
   title={{GPQA}: A Graduate-Level Google-Proof Q\&A Benchmark},
   author={David Rein and Betty Li Hou and Asa Cooper Stickland and Jackson Petty and Richard Yuanzhe Pang and Julien Dirani and Julian Michael and Samuel R. Bowman},
   booktitle={First Conference on Language Modeling},
   year={2024},
   url={https://openreview.net/forum?id=Ti67584b98}
 }

@inproceedings{
ming2024faitheval,
title={FaithEval: Can Your Language Model Stay Faithful to Context, Even If ''The Moon is Made of Marshmallows''},
author={Yifei Ming and Senthil Purushwalkam and Shrey Pandit and Zixuan Ke and Xuan-Phi Nguyen and Caiming Xiong and Shafiq Joty},
booktitle={The Thirteenth International Conference on Learning Representations},
year={2025},
url={https://openreview.net/forum?id=UeVx6L59fg}
}

@inproceedings{
Xie2024KnowledgeConflict,
title={Adaptive Chameleon  or Stubborn Sloth: Revealing the Behavior of Large Language Models in Knowledge Conflicts},
author={Jian Xie and Kai Zhang and Jiangjie Chen and Renze Lou and Yu Su},
booktitle={The Twelfth International Conference on Learning Representations},
year={2024},
url={https://openreview.net/forum?id=auKAUJZMO6}
}

@inproceedings{wang2024novelqa,
   title={Novel{QA}: Benchmarking Question Answering on Documents Exceeding 200K Tokens},
   author={Cunxiang Wang and Ruoxi Ning and Boqi Pan and Tonghui Wu and Qipeng Guo and Cheng Deng and Guangsheng Bao and Qian Wang and Yue Hang},
   booktitle={The Thirteenth International Conference on Learning Representations},
   year={2025},
  url={https://openreview.net/forum?id=uMEsKEiB7J}
 }

@misc{qwen35,
    title = {Qwen3.5: Accelerating Productivity with Native Multimodal Agents},
    url = {https://qwen.ai/blog?id=qwen3.5},
    author = {Qwen},
    month = {February},
    year = {2026}
}

@inproceedings{
qian2025scaling,
title={Scaling Large Language Model-based Multi-Agent Collaboration},
author={Chen Qian and Zihao Xie and YiFei Wang and Wei Liu and Kunlun Zhu and Hanchen Xia and Yufan Dang and Zhuoyun Du and Weize Chen and Cheng Yang and Zhiyuan Liu and Maosong Sun},
booktitle={The Thirteenth International Conference on Learning Representations},
year={2025},
url={https://openreview.net/forum?id=K3n5jPkrU6}
}

@inproceedings{
liu2024a,
title={A Dynamic {LLM}-Powered Agent Network for Task-Oriented Agent Collaboration},
author={Zijun Liu and Yanzhe Zhang and Peng Li and Yang Liu and Diyi Yang},
booktitle={First Conference on Language Modeling},
year={2024},
url={https://openreview.net/forum?id=XII0Wp1XA9}
}

@misc{meng2026equimemcalibratingsharedmemory,
      title={EquiMem: Calibrating Shared Memory in Multi-Agent Debate via Game-Theoretic Equilibrium}, 
      author={Yuqiao Meng and Sakshi Sunil Narvekar and Luoxi Tang and Rupali Rajendra Vaje and Yingxue Zhang and Muchao Ye and Zhaohan Xi},
      year={2026},
      eprint={2605.09278},
      archivePrefix={arXiv},
      primaryClass={cs.AI},
      url={https://arxiv.org/abs/2605.09278}, 
}

@inproceedings{
    tian2026multiagentdebatememorymasking,
    title={Multi-Agent Debate with Memory Masking},
    author={Hongduan Tian and Xiao Feng and Ziyuan ZHAO and XiangyuZhu and Rolan Yan and Bo Han},
    booktitle={The Fourteenth International Conference on Learning Representations},
    year={2026},
    url={https://openreview.net/forum?id=EdTt8nMAMA}
}

@misc{schulman2017proximalpolicyoptimizationalgorithms,
      title={Proximal Policy Optimization Algorithms}, 
      author={John Schulman and Filip Wolski and Prafulla Dhariwal and Alec Radford and Oleg Klimov},
      year={2017},
      eprint={1707.06347},
      archivePrefix={arXiv},
      primaryClass={cs.LG},
      url={https://arxiv.org/abs/1707.06347}, 
}

@misc{qwen3embedding,
      title={Qwen3 Embedding: Advancing Text Embedding and Reranking Through Foundation Models}, 
      author={Yanzhao Zhang and Mingxin Li and Dingkun Long and Xin Zhang and Huan Lin and Baosong Yang and Pengjun Xie and An Yang and Dayiheng Liu and Junyang Lin and Fei Huang and Jingren Zhou},
      year={2025},
      eprint={2506.05176},
      archivePrefix={arXiv},
      primaryClass={cs.CL},
      url={https://arxiv.org/abs/2506.05176}, 
}
